\documentclass[]{interact}

\usepackage{graphicx}%
\usepackage{multirow}%
\usepackage{amsmath,amssymb,amsfonts}%
\usepackage{xcolor}%
\usepackage{color,soul}%
\usepackage{amsthm}%
\usepackage[font=small]{caption}
\usepackage{subcaption}
\usepackage{mathrsfs}%
\usepackage[title]{appendix}%
\usepackage{multirow}%
\usepackage{multicol}%
\usepackage{float}%
\usepackage{manyfoot}%
\usepackage{booktabs}%
\usepackage{amsthm}%
\usepackage{array}%
\usepackage{algorithm}%
\usepackage{algorithmicx}%
\usepackage{algpseudocode}%
\usepackage{listings}%
\usepackage{ragged2e}%
\usepackage[table,xcdraw]{xcolor}
\usepackage{hyperref}
\hypersetup{colorlinks,linkcolor={red},citecolor={green},urlcolor={red}}
\usepackage[dvipsnames]{xcolor}
\theoremstyle{plain}
\newtheorem{theorem}{Theorem}[section]

\newtheorem{prob}[theorem]{Problem}
\newtheorem{assum}[theorem]{Assumption}

\theoremstyle{definition}

\theoremstyle{remark}

\begin{document}

\title{Swarm Self-Clustering for Communication-denied Environments without Global Positioning} 

\author{
\name{Sweksha Jain\textsuperscript{a}\thanks{Leena Vachhani, leena.vachhani@iitb.ac.in}, Rugved Katole\textsuperscript{b} and Leena Vachhani\textsuperscript{a}}
\affil{\textsuperscript{a}Centre for Systems and Control, Indian Institute Of Technology, Bombay; \textsuperscript{b}Computer Science and Engineering, The Ohio State University, Columbus}
}
\maketitle

\begin{abstract}
In this work, we investigate swarm self-clustering, where robots autonomously organize into spatially coherent groups using only local sensing and decision-making, without relying on external commands, global positioning, or inter-robot communication. This fully decentralized approach enables each robot to form and maintain stable clusters by responding solely to relative distances from nearby neighbors, as detected through onboard range sensors with limited fields of view. Motivated by real-world scenarios such as autonomous underwater robot retrieval and human gathering during emergency response, the proposed method is designed for GPS-denied and communication-constrained environments. Unlike conventional approaches, it requires no prior knowledge of cluster parameters such as size, number, or member identity, making it highly adaptable to unpredictable settings. A mechanism that enables each robot to adaptively alternate between consensus-based and random goal assignment based on local neighborhood size, ensuring robust, scalable, and untraceable clustering independent of initial configuration. Through extensive simulations and real-robot experiments, we demonstrate the method’s scalability, empirical convergence, and robustness under varying initial conditions and dynamic robot additions. The proposed communication-free approach outperforms local-only baselines across standard cluster quality metrics, producing clusters that exhibit untraceability even under identical initial conditions.


\end{abstract}

\begin{keywords}
Swarm Robotics, Swarm Clustering, Collective Behavior, Communication-denied Environment, Sensor-based method
\end{keywords}

\section{Introduction}
The swarm robotics paradigm draws inspiration from collective behaviors observed in various biological organisms, such as honey bees, ants, birds, and fishes \cite{brambilla2013swarm}. The collective behaviors primarily observed are flocking \cite{bahaidarah2024swarm}, clustering \cite{tan2013quorum}, aggregation \cite{8613876}, collaborative search \cite{yang2019extended}, pattern formation \cite{coppola2019provable}, and foraging \cite{9782555}. These collective swarm behaviors are typically observed in biological organisms and exhibited by utilizing easily deployable (small-size) autonomous robots. These robots like Kilobots \cite{rubenstein2012kilobot} and SWARM-BOT \cite{cheraghi2021past} have limited computational power, memory, and sensing capability with minimal environmental knowledge. In general, techniques for collective behaviors in multi-robot systems have been developed through various communication modalities that convey intent or environmental information. For instance, robots may employ local communication to exchange data with neighboring peers \cite{kengyel2016collective, rios2025robot}. In contrast, global communication refers to information exchange with all or a selected subset of robots, typically facilitated through a communication network modeled as a graph, where edges denote direct communication links. While global information sharing is also supported by technologies such as Global Positioning System (GPS), the known unreliability and signal attenuation of GPS in underwater swarm applications necessitate alternative approaches \cite{6003649}.

In this work, we address swarm clustering for two cases, where global localization and inter-robot communication are unavailable or unreliable: (i) collecting underwater robots and (ii) mimicking humans forming small groups in emergency response. In the underwater case, robots are initially aware of their global positions but rely solely on odometry within boundary limits during operation. In the emergency response case, humans move independently within their respective boundaries. The objective of self-clustering is to achieve autonomous, limited sensing-based swarm collective behavior within bounded regions, without reliance on global information or inter-robot communication. The primary challenge lies in developing swarm clustering technique using only local sensing, without any form of communication or prior knowledge of robotic attributes (e.g., shape, color), cluster properties (e.g., number of clusters, size, member identity), or predefined spatial references. Additionally, the approach must maintain low computational overhead to accommodate the limited processing capabilities of small-scale swarm robots.

In the proposed work, each robot uses only its onboard sensor to detect the positions of neighboring robots within its vicinity and independently determines its actions to form spatial clusters, without interacting with the environment or one another. This approach enables decentralized decision-making, eliminating the need for global coordination or explicit information exchange among 
robots. As a result, it circumvents common limitations associated with communication-dependent systems, such as information delays, centralized control requirements, vulnerability to malicious attacks \cite{gandhe2022decentralized}, risks of data interception or security breaches \cite{kolias2011swarm}, and the attenuation of electromagnetic signals in underwater environments \cite{6003649}. We propose a swarm self-clustering method based on assigning either consensus-driven or random, goal updates based on local neighborhood size, along with a distributed termination rule. The formed clusters are stable, compact and adaptive to any new member introduced in the swarm. We demonstrate that the swarm of robots forms a variable number of clusters with differing compositions and spatial locations across multiple runs. Consequently, both the cluster configurations and the individual robot trajectories remain untraceable from one trial to another. This implies that the evolution of cluster formation cannot be predicted or reconstructed using historical trajectory data. This key feature of untraceability highlight that the trajectories and robots in the formed cluster(s) across independent runs are inherently non-repetitive and dissimilar, thereby preventing any traceability from prior executions. The contributions of this work are summarized as follows:

\begin{itemize}
    \item A novel decentralized consensus-inspired method is proposed that uses only local sensing, without inter-robot communication or global localization, to form compact and stable clusters without prior knowledge of cluster parameters such as number of clusters, cluster size, or member identities.
    \item Comprehensive quantitative and qualitative comparisons with local-only baseline methods are conducted using standard cluster quality metrics like compactness, silhouette coefficient, and Dunn index along with physical metrics such as computational complexity and convergence time. The combined evaluation of these measures demonstrates the overall clustering efficiency.    
    \item The scope of this method claims empirical convergence, scalability to large swarm sizes, and adaptability to expansion on new arrivals with extensive simulations and real-world experiments. Method’s untraceability property is evaluated through appropriate performance metrics.       
\end{itemize}
This paper is organized as follows: the next section finds the research gap in relevant literature and Section \ref{CF} formulates the problem. The detailed methodology for the proposed solution with analytical results is presented in Section \ref{SSA}. Section \ref{ES} provides simulation, experimental results, corresponding discussions and analysis. This section also entails comprehensive comparative results with the state-of-the-art methods. Section \ref{C} concludes with the features of the proposed swarm self-clustering method and future work. Additional results and representative scenarios with supplementary material are provided in Appendix.

\section{Previous Work}
In the existing literature, various swarm algorithms have been developed to exhibit a wide range of collective behaviors, including aggregation, flocking, clustering, pattern formation, and many others. This section outlines prior works on those specific behaviors that closely resemble the swarm dynamics addressed in our proposed approach. Since the proposed work primarily exhibits swarm clustering, several aspects of the method and its key features also align with characteristics observed in aggregation and flocking behaviors. In the collective behavior of aggregation, a group of robots maintain spatial proximity, forming a compact formation without requiring explicit communication or coordination mechanisms \cite{gauci2014self}. In flocking, robots not only maintain proximity but also align their headings and velocities using consensus driven approach, enabling coherent group movement in a common direction \cite{HENGSTEBECK2024100160}. However, swarm clustering is the process by which robots self-organize into multiple spatially distinct groups, with each group exhibiting local proximity and cohesion without relying on predefined cluster locations or global information \cite{cruz2017robust}.

Existing approaches to swarm aggregation are broadly classified into two categories: 1) those that rely on external environmental cues \cite{na2021bio, rios2025robot}, and 2) those that are self-organized, emerging from local interactions among robots without environmental dependency \cite{BAKHSHIPOUR2017708}. The cue-based swarm aggregation \cite{brambilla2013swarm}, uses cues such as light, heat, or sound sources, while self-organized aggregation considers the interaction of robots with the environment and each other to achieve swarm aggregation \cite{correll2011modeling}. The collective behavior wherein such aggregates move cohesively in the same direction is flocking. 
Flocking has been achieved through the use of boids (bird-old object), the autonomous robots \cite{hartman2006autonomous} that follow three fundamental rules: cohesion, separation, and alignment. The framework preserves its computational simplicity with the addition of a fourth rule, local leader following, which improves behavioral fidelity by modeling the coordinated motion characteristic of biological flocks. The boids has also been extended to address safety-critical environments \cite{HENGSTEBECK2024100160}, however it needs changes in the environment to distinguish environment boundary. The flocking behavior has also been implemented using Particle Swarm Optimization (PSO) \cite{bahaidarah2024swarm}, wherein each agent iteratively updates its position and velocity based on both personal and neighborhood bests, resulting in emergent flocking behavior. Both of these methods Boids and PSO implements consensus-based swarm models to achieve coordinated motion through deterministic attraction toward a shared state or virtual leader. As observed in flocking \cite{olfati2007consensus} the convergence of these methods depends on global or local communication models. A self-organized flocking in 3D \cite{karaguzel2024self} has onboard implementation where the drones communicates using a radio communication channel. Techniques that do not use global information, share any information/intent and independently make decisions to achieve collective behavior is worth investigating.

The proposed work investigates swarm self-clustering, a collective behavior wherein each robot independently aims for forming cluster and as a result, spatially distinct cluster(s) emerge without centralized control and any global information. {Unlike the consensus based methods, we propose a consensus-based switching method based on assigning consensus-driven or random goal updates based on local neighborhood size, along with a distributed termination rule to converge by forming stable clusters.} There is no a priori specification of the number of clusters formed. Among the various collective behaviors studied in swarm robotics, clustering has particularly drawn inspiration from the self-organizing capabilities observed in natural systems, such as ant colonies and termite aggregations \cite{BAYINDIR2016292}. The Beeclust algorithm \cite{5952587}, a swarm clustering method enables robots to form clusters by leveraging their local sensing capabilities. In this method, goal locations are implicitly defined through environmental cues, allowing robots to identify and aggregate around them using simple sensor-based rules. Other clustering techniques rely on similarities in features such as size, shape, or observable properties—for instance, robots blinking the same color LED light—to facilitate grouping \cite{cruz2017robust}. These strategies often aim to optimize task allocation or resource distribution among the swarm.

The swarm clustering methods predominantly is categorized into two types: (i) Token clustering, and (ii) Robot clustering. In token clustering, robots create clusters of objects/tokens with similar features (such as color, size or shape) \cite{aguzzi2023field}, and involves both homogeneous and heterogeneous groups of robots. On the other hand, in robot clustering, robots must move from their initial positions to form clusters when the number of clusters is known. The robots form clusters based on the specified number of clusters and the positions of their respective centroids, typically involving homogeneous robots \cite{mccune2014decentralized}. The clustering algorithms like k-means clustering and DBSCAN (density-based spatial clustering of applications with noise) \cite{deng2020dbscan, anikin2017privacy} aim for spatial clustering, not necessarily aimed at robotic swarms. Furthermore, segregation is a collective behavior where different spatial patterns of similar robots for a heterogeneous swarm of robots is observed \cite{kumar2010segregation, santos2020spatial}.

Further, swarm robotics methods can be categorized by their implementation approach \cite{brambilla2013swarm}, which primarily includes virtual physics-based, probabilistic, and evolutionary techniques for achieving collective behaviors. In Virtual Physics-based approaches \cite{khaldi2018self}, robots interact based on virtual forces inspired by physical laws. They experience attractive forces that pull them toward a goal while repulsive forces push them away from obstacles, ensuring effective navigation and collision avoidance. Probabilistic approaches \cite{4937849}, on the other hand, model robot dynamics using Markov chains, allowing the system to predict aggregation behaviors based on transition probabilities. This method helps estimate the likelihood of robots joining or leaving a swarm formation, making it particularly useful for self-organizing systems. Lastly, evolutionary approaches \cite{yang2019extended, 8594056, 5952587} leverage artificial evolution to optimize swarm behavior over time. There are methods like \cite{duarte2016evolution,1501639}, where evolved collective behavior has been showcased. In the work by Dorigo et al. \cite{dorigo2004evolving}, the evolution of controller is presented to showcase swarm aggregation. The sound sources present in the environment work as cue and help in building communication among s-bots. While the probabilistic aggregation strategy presented by Soysal and Sahin (2005) \cite{1501639} specifically deny the portability of the controllers onto the physical robots. 

\begin{table*}[htbp]
\centering
\caption{Literature summary}
\label{table}
\begin{center}{
\scalebox{0.75}{
\begin{tabular}{|m{4.5em} | m{10em} | m{1.7cm}| m{1.7cm} | m{1.5cm} | m{1.7cm} | m{3.5cm}|} 
 \hline
 Collective Behavior & Algorithm & Decentral-ized & Communi-cation free & Local Sensing & Global Positioning & Known information \\
 \hline
 \multirow{6}{2cm}{\textbf{Swarm Aggrega-tion}} & Cue-based aggregation \cite{na2021bio} & Yes & No & Yes & No & Virtual pheromones\\
    \cline{2-7}
    & Beeclust algorithm \cite{rios2025robot} & Yes & Yes & Yes & No & Area of aggregation\\
    \cline{2-7}
  & Population Dynamics Model \cite{correll2011modeling} &  Yes & No & Yes & No & Communication sensor\\
  \cline{2-7}
  & Search and Rescue \cite{BAKHSHIPOUR2017708} & Yes & No & Yes & Yes & Victim's location is global best\\
  \cline{2-7}
  & Fuzzy Based Method \cite{misir2020fuzzy} & Yes & No & Yes & Yes & Fuzzy rules\\
  \cline{2-7}
  & DW-KNN topology \cite{khaldi2018self} & Yes & No & Yes & Yes & Target position \\
   \hline
 \multirow{3}{2cm}{\textbf{Flocking}} & Autonomous Boids \cite{hartman2006autonomous, HENGSTEBECK2024100160} & Yes & Yes & Yes & No &Local leaders /Ghost boids\\
 \cline{2-7}
  & Particle Swarm Optimization \cite{bahaidarah2024swarm} & No & No & Yes & Yes &Local and global best position \\
  \cline{2-7}
  & Self-organized Flocking \cite{karaguzel2024self} & Yes & No & No& Yes & Radio communication channel\\
 \hline
\multirow{5}{2cm}{\textbf{Swarm Clustering}} & Spatial Clustering\cite{cruz2017robust} & No & No & Yes & Yes  & Color of target cluster\\
  \cline{2-7}
  & DBSCAN\cite{deng2020dbscan, anikin2017privacy} & Yes & No & Yes & Yes &  Cluster density\\
  \cline{2-7}
  & K-means Clustering \cite{mccune2014decentralized} & Yes & No & No& Yes & Number of Clusters \\
  \cline{2-7}
  & Segregation \cite{kumar2010segregation, santos2020spatial} & Yes & Yes & Yes & Yes & Spatial patterns\\
  \cline{2-7}
  & \textbf{Proposed Swarm Self-Clustering} & \textbf{Yes} & \textbf{Yes} & \textbf{Yes} & \textbf{No} & \textbf{Boundary limits}\\
 \hline
\end{tabular}}}
\end{center}
\end{table*}

Table~\ref{table} summarizes existing work to highlight the distinguishing features of the proposed approach. While certain methodologies, such as Beeclust, share conceptual similarities with the proposed approach, the parameters and features used to achieve the target behavior differ significantly which improves the behavior of the proposed method drastically. Other approaches like the Boids class of algorithms for flocking, also emphasizes independent decision making, but rely on ghost robots to avoid collisions with the environment boundaries, an approach that requires either modifications to the environment or the introduction of dummy robots. The proposed method extend the classical consensus based approach as observed in Boids and PSO to a consensus inspired method with adaptive alternating goal updates mechanism for swarm clustering. This approach thereby helps in forming robust and untraceable clusters because of alternate goal assignments along with distributed termination rule. The technical novelty is further highlighted in \textit{Goal Assignment section \ref{section 2}}.
 
This work is motivated by the objective to explore decentralized collective behaviors emerging from locally perceived information, with robots acting autonomously without any reliance on explicit communication. Several communication strategies have been introduced to reduce reliance on GPS data in underwater swarm applications; however, the success rates of these approaches suggest ample room for improvement.
To the best of our knowledge, no prior work has addressed swarm clustering in communication-denied environments using only range-only sensing to achieve self-organized, emergent collective behavior.

\section{Problem Formulation}\label{CF}
Given the robots with onboard capabilities of only range sensing and constrained computing, the objective of swarm self-clustering is to form one or more clusters of robots. In this work, a cluster is defined as {\em a group of at least $M$ spatially close robots,} where $M$ is the lower bound on cluster size. Each robot makes independent decisions, does not communicate, and is equipped with a range sensor which measures the distances and angles of the detected robots in the sensor's limited field of view (FoV). In the absence of global localization, each robot operates within a bounded region that constrains its motion during exploration. These bounds act only as motion constraints for random goal generation and do not require robots to possess any knowledge of global boundaries or positional coordinates. To ensure that robots are not spatially isolated by construction, we assume that the operational regions associated with the robots at the initial time satisfy a geometric connectivity condition. This condition is formally stated in Assumption~\ref{assum1} and forms the basis for the expected convergence behavior of the proposed method.

\begin{assum}
Let $\mathcal{B}_i(0) \subset \mathbb{R}^2$ denote the bounded region associated with robot $i$ at the initial time. We assume that the union of these regions forms a connected set, i.e.,
\begin{equation}
\bigcup_{i=1}^{S} \mathcal{B}_i(0) \text{ is connected}.\nonumber 
\end{equation}
\label{assum1}
\end{assum}
\vspace{-0.5cm}
The assumption above ensures that robots are not spatially isolated at initialization. Under bounded random exploration, robots eventually move through the connected environment and can detect nearby neighbors using their range sensors. This assumption is imposed only on the initial deployment and does not require robots to possess global localization, shared boundary information, or any knowledge of the regions associated with other robots.
\begin{figure*}[htbp]
    \centering
    \includegraphics[width = 4 in]{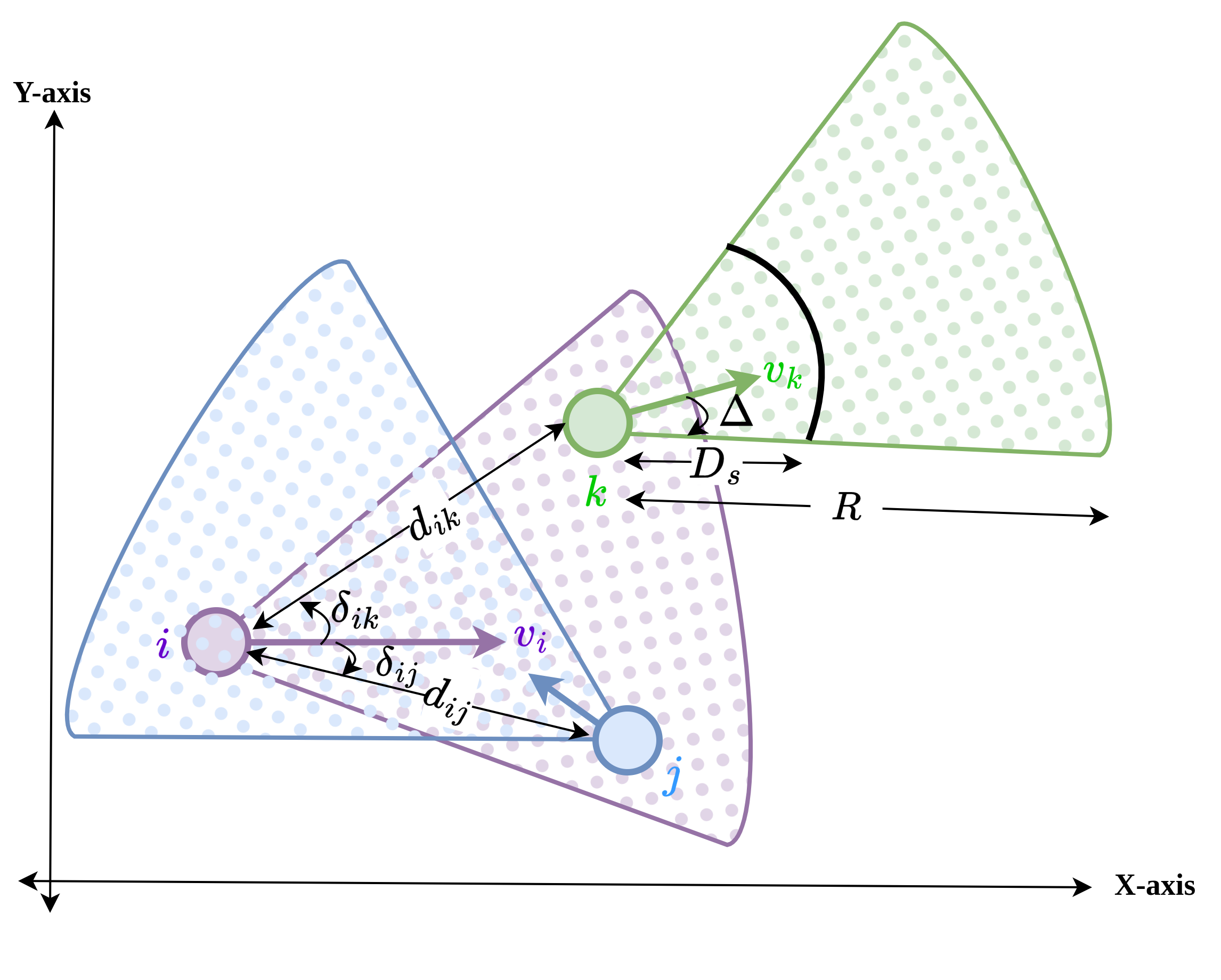}
    \caption{Illustration of inter-robot detection, where $i, j, k$ represents the 2D robots with conic/restricted FoV with sensing range $R$, angle $2\Delta$, and a safe distance $D_s$. The sensed distance and bearing of robot $i$ with respect to robot $j$ are $d_{ij}$ and $\delta_{ij}$, respectively.}
    \label{CA}
\end{figure*}
Each robot detects its neighbors within its sensor's FoV. Figure \ref{CA} depicts a scenario with three robots: $i$, $j$, and $k$. Each robot considers limited sensing FoV (conic) with angle $2\Delta$, and sensing range $R$. A safe distance, indicated by $D_s$, is the minimum distance between two robots to avoid inter-robot collisions. Figure \ref{CA} illustrates a scenario where the distance of robot $i$ concerning robot $j$ and $k$ is less than the sensing range. The robots $j$ and $k$ lie in the FoV of robot $i$; hence, both are detected/sensed by robot $i$. Robot $k$ and robot $j$ do not lie within each other's field of view (FoV), and therefore, no detection occurs. However, robot $i$ is in FoV of robot $j$; hence, robot $i$ gets detected by robot $j$. Further, the command inputs to the robot $i$ are linear velocity $v_i$ and angular velocity $\omega_i$. Now, formally the problem of swarm self-clustering is formulated as follows:
\textbf{
\begin{prob}\label{pro}
    Devise a simple and computationally efficient communication-free decentralized swarm algorithm for robots with only limited range sensing capability, where swarming robots must form cluster(s) based on local proximity and motion within bounded regions, without access to global localization and any additional information about the robots or the environment.
\end{prob}}
It is important to note that a key challenge in developing a self-clustering method lies in ensuring that each robot successfully integrates into a spatially compact cluster while concurrently maintaining the integrity of existing cluster formations. Table~\ref{terms} lists all the terms and their definitions used in the manuscript.

\begin{table*}[h!]
\centering
\caption{Glossary of terms}
\renewcommand{\arraystretch}{1.3}
\begin{tabular}{@{}>{\bfseries}l l p{8cm}@{}}
\toprule
\textbf{Name} & \textbf{Term} & \textbf{Definition} \\ \midrule
Robots & $i, j, k$ & Index used to represent individual robots in the swarm \\
Sensing range & $R$ & Maximum distance within which a robot can detect obstacles or neighbors using sensors \\
Field of View (FoV) & $2\Delta$ & Angular range within which a robot can perceive its surroundings \\
Swarm size & $S$ & Total number of robots in the swarm \\
Minimum cluster size & $M$ & Minimum number of robots required to form a valid cluster \\
Neighbor set & ${N}_i$ & Set of all robots within the neighborhood of robot $i$ \\
Size of neighborhood & $|N_i|$ & Number of robots in the neighbor set of robot $i$ \\
Goal Distance & $D_g$ & The maximum radial distance between a robot and its assigned goal for the robot to stop\\
Safe Distance & $D_s$ & Minimum inter-robot separation to avoid collision\\
Number of clusters & $K$ & Number of distinct robot clusters formed in the environment \\
\bottomrule
\end{tabular}
\label{terms}
\end{table*}

\section{Methodology}\label{SSA}
The proposed approach establishes a fully decentralized methodology that is executed independently by each robot, enabling the emergence of self-organizing swarm clustering behavior. Such behavior is governed by four key parameters: 1) the neighbor set $N_i$ of robot $i$,  2) minimum cluster size $M$, 3) goal distance $D_g$, and 4) safe distance $D_s$. The definitions and roles of these parameters are introduced and explained in the corresponding context.

\vspace{0.2cm}
\noindent \textbf{\textit{Neighbor Set}}

\noindent The neighbor set is given by 
\begin{align}
    N_{i} = \lbrace j~|~ d_{ij} < R ; \delta_{ij} < \Delta \rbrace,
\end{align}
where $d_{ij}$ is the inter-robot distance between the robot $i$ and the robot $j$, as measured by the range sensor onboard the robot $i$. 

\vspace{0.2cm}
\noindent \textbf{\textit{Minimum Cluster Size}}

\noindent Given the minimum cluster size $M$, a cluster is formed when each robot has number of neighbors more than or equal to $M-1$. Therefore, the objective of self-clustering becomes to ensure that cardinality of the neighbor set $|N_i|$ is at least $M-1$ for each robot $i$. Further, the minimum cluster size $M$ sets an upper bound on the total number of clusters formed for the swarm size of $S$, given by
\begin{align}\label{upperbound}
    \text{Number of clusters} \leq \left\lfloor\frac{S}{M}\right\rfloor,
\end{align}    
where $\lfloor \cdot \rfloor$ gives the greatest integer less than or equal to the argument. The minimum cluster size can be used to form a single cluster by initializing it to satisfy $M \geq S/2+1$.

A straightforward strategy for cluster formation involves assigning a common goal to a group of neighboring robots and navigating them toward that goal while avoiding collisions. However, in the absence of global localization and inter-robot communication, maintaining a shared neighbor set and assigning a common goal becomes challenging. To address this, we propose a self-organizing swarm clustering framework achieved through three fundamental and continuously executed tasks by each robot:
(i) {\em Neighbor Detection}, (ii) {\em Navigation}, and (iii) {\em Goal Assignment}.

\subsection{Neighbor Detection} \label{neigh_detect}
After initialization, each robot starts detecting its neighbors instantly. This task helps each robot to form its neighbor set. A robot is detected in other robots' FoV via an onboard range sensor. 
All the following conditions (in line with the range sensing) should be satisfied for the robot $j$ to be a neighbor of the robot $i$:
\begin{itemize}
    \item Distance between two robots is less than the sensing range of the robot, i.e., $d_{ij} < R$, and;
    \item The bearing of the robot $j$ with respect to the robot $i$ is less than the half of FoV of the robot $i$ i.e. $\delta_{ij} < \Delta$.
    \item The robot $j$ is in Line-of-Site (LoS) of robot $i$ (if robot $j$ is occluded by any other robot $k$, then robot $k$ is a neighbor of the robot $i$).  
\end{itemize}

\noindent \textbf{Remark}: If a robot $j$ is detected by the robot $i$, then $j$ will belong to $N_{i}$ but the converse is not always true unless $\Delta = \pi$.

When robot $i$ detects robot $j$, the cardinality of the neighbor set increases to $|N_i|+1$. Subsequently, the robot $i$ estimates the positions of neighboring robots by utilizing range sensor readings $d_{ij}$ and $\delta_{ij}$ with respect to itself (in it's local frame). As the robot $i$ lacks global information, it consistently assumes its own location at the origin, such that $\{x_i, y_i\} = \{0, 0\}$. The location of the detected robot $j$, is computed by:

\begin{equation}
\begin{aligned}
    x_j =  d_{ij}cos(\delta_{ij})
    ;~y_j =  d_{ij}sin(\delta_{ij}) 
\end{aligned}    
\label{eq:robot}
\end{equation}
Hence, the relative location of each robot $j$ in the neighbor set $N_i$ becomes available to the robot $i$ for goal assignment task and collision avoidance utilized in navigation task. It is important to note that the neighbor set of each robot comprises the estimated positions of its neighboring robots, and hence these measurements are inherently subject to sensor noise and estimation uncertainty. Each robot $i$ maintains its neighbor set $N_i$ and gives an unique identification to each robot in its neighbor set $N_i$. This unique identification is local to each robot.

\subsection{Navigation}\label{Nav}
The navigation controller commands each robot to navigate towards their respective goal position, avoiding collision with other robots. The underlying objective of every robot's controller is to reach its assigned goal while avoiding the collision with other robots. A navigation controller assigns linear and angular velocities to each robot, resulting in the movement of robots towards their assigned goal. The navigation controller ensures a collision avoidance by keeping a safe distance $D_s$ with other robot, where $D_s$, is the minimum distance between two robots to avoid inter-robot collisions. In this work, we use a navigation controller (taken from \cite{8613876}), that ensures collision-free motion to reach the goal. However, cluster formation (i.e., reaching the assigned goal) and collision avoidance can become conflicting objectives, particularly when two robots have goals located in close proximity. To address this, we introduce the parameter $D_g$, a predefined threshold representing the radial distance between a robot and its instantaneous goal position. A robot is considered to have reached its goal when it is within a distance $D_g$ of it. The parameters $D_s$ (safe distance) and $D_g$ (goal distance) are both predefined design parameters, and the rationale behind their chosen values is discussed in the following section.
\vspace{0.5 cm}

\noindent \textbf{\textit{Lemma 1:}}
A necessary condition to achieve synergy is  $D_s \leq D_g$.

\noindent \textit{Proof}: To achieve the swarm self-clustering, the proposed method is based on assigning a goal such that each robot must stop once it is close enough to other robots. The safe distance $D_s$ ensures collision avoidance by maintaining a minimum inter-robot separation, while the goal distance $D_g$ determines the threshold for goal attainment. A robot successfully completes ``reaching the assigned goal" task, only when $D_g > D_s$. If this condition is not satisfied, the robot will prioritize collision avoidance over goal convergence, thereby preventing cluster formation.
$\blacksquare$

\subsection{Goal Assignment}\label{section 2}
The goal assignment for each robot is based on the available sensor-based estimation of neighbors' locations. To bring the robots closer to each other, we propose a basic goal update rule (consensus-based), which is finding the centroid of neighbor set for each robot. The goal position $\{x_{g_i}, y_{g_i}\}$ of robot $i$, calculated using goal update rule is given by:
\begin{equation}
    \begin{aligned}
        x_{g_i} = \frac{(\Sigma_{j \in N_{i}} x_{j})}{|N_i|+1}; 
        ~y_{g_i} = \frac{(\Sigma_{j \in N_{i}} y_{j})}{|N_i|+1}
        \label{eq:tar}
    \end{aligned}
    \end{equation}
Where ($x_{j}, y_j$) are the position coordinates of robot $j$ in neighbor set $N_i$. The proposed approach assigns a goal to the robot $i$ based on its corresponding neighbor set $N_i$. The goal assignment mechanism plays a crucial role in guiding robots to form clusters by adaptively alternating between {\em consensus-based and random goal selection}, depending on each robot’s local neighborhood configuration. The robot trajectories are bounded because of the random goal assignment within the boundary. The goal assignment is triggered under two conditions. The first occurs when the robot reaches its current goal within the threshold distance $D_g$ (where $2D_g$ represents the maximum inter-robot separation between two robots in a cluster). The second condition arises when there is a change in the number of neighbors. Robots that are already part of an existing cluster will move only if their instantaneous distance to the assigned goal, $d_g$, exceeds $D_g$. Accordingly, the {\em Goal Assignment} task is governed by the three conditions summarized in \eqref{goal}. 
\begin{equation}
\begin{array}{cc}
    \text{Goal}  \\
    \text{Assignment} 
\end{array} =~    
    \begin{cases}
        \text{Random}, & \text{if } d_g \leq D_g~\&~|N_i|=\phi\\
        \text{Goal~Update~Rule}, & \text{if } d_{ij} \leq R ~\&~|N_i|~\text{changes}\\
        \text{Does~not~change}, & \text{otherwise}\\
    \end{cases}\\
    \label{goal}
\end{equation}

While individual components such as consensus-based updates (centroid goal updates) or random exploration have been extensively explored in prior work, the improvements reported in this work arise from their coordinated use through adaptive goal selection and a termination mechanism.\\

\noindent \textbf{Algorithm}

\begin{figure}[!h]
    \centering
    \includegraphics[width=0.9\linewidth]{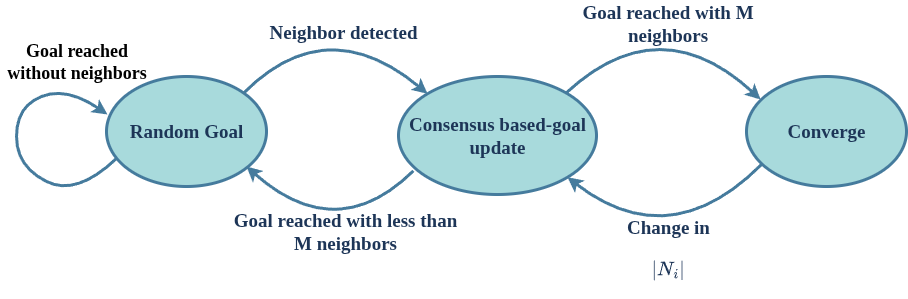}
    \caption{State-machine diagram explaining adaptive goal-switching logic}
    \label{fig:state}
\end{figure}

\noindent The pseudo-code for the proposed algorithm is presented in Algorithm \ref{pseudo_code1}. All the robots execute the same algorithm independently to form one or more clusters. The algorithm is designed to work in an infinite loop to adapt any changes in the swarm. The robot $i$ for each $i=1,2,\ldots S$ continuously searches for neighbors using the range sensor (line 2) and starts detecting neighbors. When the robot $i$ has zero neighbors, a uniformly distributed random goal within the limit of goal update boundary is assigned to the robot $i$ (lines 4-5). Otherwise, the goal is updated using the goal update rule (lines 7-8). Then, a navigation controller assigns the robot the linear and angular velocity to move towards its assigned goal (lines 20-24), which is known to the robot $i$ only. During this process, if the robot $i$ successfully achieves its goal while adhering to the specified minimum cluster size $M$, its goal is adjusted to its current position (lines 12-13) making it stop to form a cluster. However, if the robot attains its goal with fewer neighbors than required, it is assigned a random goal again (lines 14-16) to allow the robot to explore forming/joining a cluster. The proposed algorithm \textit{converges} when each robot is a member of a cluster and each robot has reached a state with goal position as its current location (line 21). To summarize, the conditions to {\em form clusters} are: 1) number of neighbors are at least $M$ and 2) Distance to the goal is less than or equal to $D_g$. Equation \eqref{goal} together with the convergence condition specified in Algorithm \ref{pseudo_code1} (lines 11–13), encapsulates the core technical contribution of the proposed method. Figure \ref{fig:state} illustrates the goal-assignment logic as a finite-state machine to clarify the relationship between Eq.\ref{goal} and Algorithm \ref{pseudo_code1}.\\

\begin{algorithm}[!h]
\textbf{Input} $R,~D_g,~\Delta, M,~D_s $ \\
\textbf{Output} $v_i,\omega_i$
\begin{algorithmic}[1]
\While {\text{Cluster not formed}} \\
$N_i$=Detect\_Neighbors($R, \Delta$)
\If{$|N_i| = \phi$}    \Comment{No Neighbor}
    \State $x_{g_i} \gets rand(boundary\; limits)$
    \State $y_{g_i} \gets rand(boundary\; limits)$
\Else    \Comment{Neighbor is present}
    \State $ x_{g_i} \gets \frac{\Sigma_{j  \in N_{i}}(x_j) }{|N_{i}|+1}$
    \State $ y_{g_i} \gets \frac{\Sigma_{j \in N_{i}}(y_j) }{|N_{i}|+1}$\\
\State $d_g = \sqrt{x_{g_i}^2 + y_{g_i}^2}$ \Comment{distance to goal}
    \If {$d_g \leq D_g$} \Comment{Target is reached}
        \If{$|N_i|+1 \geq \textit{M}$}\Comment{Minimum cluster size condition is satisfied}
            \State $\lbrace x_{g_i},y_{g_i} \rbrace= \lbrace 0,0 \rbrace$
        \Else 
            \State $x_{g_i} \gets rand(boundary\; limits)$
            \State $y_{g_i} \gets rand(boundary\; limits)$
         \EndIf
    \EndIf
\EndIf
\If {$\lbrace x_{g_i}, y_{g_i}\rbrace = \lbrace 0,0 \rbrace$}
    \State $\lbrace v_i, \omega_i \rbrace $ = $\lbrace 0,0 \rbrace$
\Else
    \State $\lbrace v_i, \omega_i \rbrace $ = \text{Navigation\_Controller}$(x_{g_i}, y_{g_i})$
\EndIf
\EndWhile
\caption{Proposed Algorithm for Robot $i$}
\label{pseudo_code1}
\end{algorithmic}
\end{algorithm}

\noindent \textbf{Remark:} The random goal for each robot is generated within a bounded region that constrains its motion during random goal assignment.  Regardless of whether the environment has a physical boundary or is open—as in underwater scenarios—this mechanism ensures that each robot’s motion remains within its bounded area. Uniform sampling within symmetric boundaries yields zero-mean expected motion updates, ensuring that odometry drift remains bounded over time. Although boundaries may drift, the swarm is {\em expected to converge} provided that robots’ boundary regions overlap sufficiently, given Assumption \ref{assum1}. Under this condition, random exploration enables inter-robot encounters, after which adaptive goal update logic drives self-clustering.\\ 

\noindent  \textbf{\textit{Compact Cluster}}\\
A compact cluster is characterized by robots that are non-collinearly aligned while maintaining close proximity to one another. The robots within such a cluster preserve this proximity by maintaining a predefined safe inter-separation distance $D_s$, thereby avoiding collisions while ensuring cohesive formation. The resulting compactness of the clusters is further evaluated and discussed in Section~\ref{ES} under the cluster quality metrics evaluation.
Let us consider the minimal case of cluster formation with the smallest possible number of robots, i.e., $M=2$. . Since the proposed algorithm ensures that every robot must be part of a cluster with at least one neighbor within its field of view (FoV), the case $M = 1$ is not feasible. Hence, a valid cluster consists of a group of two or more robots ($M \ge 2$). For a cluster of two robots $M=2$ , the maximum inter-robot distance at steady state is $2D_g$ and the minimum required sensing range must satisfy $R > 2D_g$ to ensure mutual visibility within their FoVs. However, in this configuration, non-collinearity and compactness cannot be guaranteed, as only two points can define a line, not a compact formation. Therefore, a compact cluster can only emerge when $M>2$. The sufficient condition for forming compact clusters is as follows:\vspace{0.4cm}

\noindent \textbf{\textit{Lemma 2:}}\\ 
A sufficient condition for the formation of a compact, non-collinear cluster of robots is that the sensing range 
$R$ satisfies $R \ge 4D_g$, where $D_g$ denotes the goal distance.\vspace{0.3cm}

\begin{figure}
    \centering
    \includegraphics[width = 3.5in]{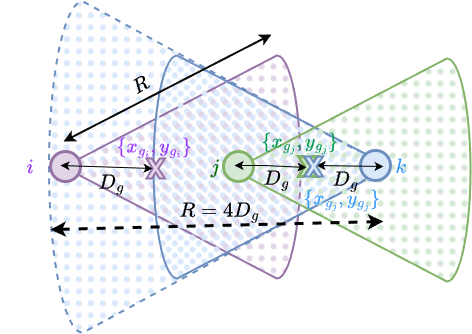}
    \caption{Collinear robots forming a cluster; small circles represent robot positions, cone represents their FoV, and 'x' mark represents the goal position.}
    \label{collinearity}
\end{figure}

\noindent \textit{Proof}: 
We use contradiction to prove and a corresponding scenario is illustrated in Figure~\ref{collinearity} for the case of $M=3$. In Figure~\ref{collinearity}, three robots $i$, $j$, and $k$ can be seen aligned to form the non-compact cluster (aligned in a line). This happens when robot $j$ gets detected by both the robots $i$ and $k$, but robot $j$ only detects robot $k$. According to step 7 of the proposed Algorithm~\ref{pseudo_code1}, goal of the robots $j$ and $k$ gets updated by using positions of each other, while the goal of robot $i$ is updated based on the position of $j$. A robot stops when it is $D_g$ distance away from its instantaneous goal and minimum $M$ robots are in the neighborhood. In the scenario with cluster of three robots aligned collinear, the robots $i$, $j$, and $k$ would stop at positions aligned along the line joining their respective goal positions. But, if $R \geq 4D_g$, at least one robot (robot $k$ in Figure~\ref{collinearity}) detects two other robots, which updates the neighbor set and subsequently its goal position. 

Under the same collinear case, for the robot $k$, the robot $j$ occludes the robot $i$. This indicates that robot $k$ has $|N_k|=1$ for $M=3$, hence robot $k$ would not stop to form cluster. In case of no occlusion, at least one robot will see two other robots, guaranteeing its goal location to not be in the line with other robots (refer \eqref{eq:tar} and use $|N_k|=2$). Hence, in this way three robots can not be collinear, if $R \ge 4D_g$, while forming cluster. $\blacksquare$\vspace{0.3cm}

\noindent \textbf{\textit{Theorem 1:}} In each formed cluster, the inter-robot distance between any two neighboring robots is bounded within the interval $(D_s, 2D_g)$, where $D_s$ is the predefined safe distance for collision avoidance, and $D_g$ is the predefined radial distance with respect to the assigned goal position. The lower bound ensures safety, and the upper bound ensures compactness.

\noindent \textit{Proof:} By design, each robot maintains a minimum distance of at least $D_s$ from all neighboring robots during motion. This safe distance is enforced through collision avoidance mechanism by the navigation controller, ensuring that robots do not come into close proximity, thereby satisfying the lower bound $D_s$. 

In a compact cluster, no three robots should be linearly aligned. From Lemma 2, it is established that the minimum sensing range must satisfy $4D_g$ to enable the formation of compact clusters. This condition inherently ensures that the robots within a cluster are non-collinear. Considering a subset of three robots, if two robots are separated by a distance of $2D_g$, the third robot must necessarily be located at a distance less than $2D_g$ from at least one of them to form a valid cluster. The only scenario that violates this condition is when the third robot lies collinearly with the other two, thereby obstructing compact formation. Hence, this reasoning establishes the upper bound on inter-robot distance necessary for achieving compact cluster formation. $\blacksquare$

\section{Results and Discussion}\label{ES}
The simulations were performed on a 12th Gen Intel® Core™ i7-12700F × 20 processor using the Robot Operating System (ROS) Noetic and Gazebo, running on Ubuntu 20.04 LTS. The robot used for simulation and real experiments is Turtle-bot3, a differential drive robot equipped with a 2D LIDAR. We spawned swarm of robots with varying swarm sizes in simulation and real environment to analyze the performance of the proposed method. 

All robots were initialized at fixed positions using the same Gazebo ROS launch file across all trials and algorithms, ensuring identical initial configurations and environments. Randomness in the experiments arises solely from algorithmic components such as random goal assignment and sensor noise. Random seeds were not explicitly fixed; instead, each trial relied on a uniform distribution using NumPy’s default pseudo-random number generator, resulting in independent realizations. All reported results correspond to statistical summaries (mean and standard deviation) computed over 30 independent runs, thereby capturing robustness with same initial conditions. The navigation controller used for implementation, the performance metrics, features and comparative analysis are presented in the following sections. 

\subsection{Navigation Controller}
The inputs to the robot i.e. linear and angular velocity applied will vary on the basis of the number of neighbors in the robots' FoV and the instantaneous distance between the robot and its neighbors. The \textit{Goal Assignment} task considers the presence of neighbors and assigns a navigation goal to the robot. The navigation controller used for implementing the algorithm is taken from Shah and Vachhani [2019] \cite{8613876}. The linear velocity $v_i$ and angular velocity $\omega_i$ of the robot $i$ are given by \eqref{eqn:kin_swarm1} and \eqref{eqn:kin_swarm2} respectively. \begin{equation}
\begin{aligned}
    v_i = 
    \begin{cases}
        v_{max}, & \text{if }d_{ij} > D_s\\
        max(v_i(t^i_e)- (t - t^i_e)\lambda_i,0),& \text{otherwise }\\
    \end{cases}\\
        \label{eqn:kin_swarm1}
\end{aligned}    
\end{equation}
\begin{equation}
    \omega_i = 
    \begin{cases}
        a_i sgn(\Dot{\theta_i}),& \text{if }d_{ij} > D_s \\
        \beta_i sgn(\Dot{\theta_i}) - b_i sgn(\delta_{ij}),& \text{otherwise },\\        
    \end{cases}
    \label{eqn:kin_swarm2}
\end{equation}

\noindent Where $t^i_e$ is the time at which the robot $i$ detects a neighbor within the safe distance, $v_i(t^i_e)$ is the linear velocity of the robot at the time when it detects a neighbor within the safe distance, and $t$ is the total simulation time, while  $\beta_i$, $b_i$, $a_i$, and $\lambda_i$ are controller's constant parameters. The parameters values related to robot and controller used for the implementation (same for each $i=1,\ldots, S$) are given in Table \ref{table:1}.

The controller presents different control laws for the swarm of robots to execute in two different cases based on the inter-robot distance with respect to the safe distance $D_s$.  If the inter-robot distance is greater than safe distance $D_s$, then the controller constantly tries to reach to the goal by aligning its orientation towards the goal using the bearing angle $\theta_i$.The bearing angle $\theta_i$ is the angle between the line connecting the current position of robot $i$ and its goal position $(x_{g_i}, y_{g_i})$, and the current orientation of robot $i$. In the other case, the controller minimizes the linear velocity and changes the orientation of the robot in a way that it can avoid collision with its neighbor, where inter-robot distance is less than $D_s$. The Algorithm~\ref{pseudo_code1} is executed on each robot, stopping it when it is within a distance $D_g$ of its assigned goal. 

\begin{table}[htbp]
\centering
\caption{Variables and their values}
\begin{tabular}{|c | c c c c c c c c|} 
 \hline
 Parameters & $v_{max} (m/s)$ & $a$ & $\beta$ & $b$ &  $D_g (m)$ & $\delta_c$ & $D_s (m)$ & $\lambda_i$ \\ [0.5ex] 
 \hline
 Values & 0.22 & 0.3 & 0.3 & 0.866 & 0.875 & $60^\circ$ & $0.775$ & $10^{-5}$ \\ [1ex] 
 \hline
\end{tabular}
\label{table:1}
\end{table}

\subsection{Simulation and Experimental Results}
\begin{figure*}[!h]
    \centering
    \includegraphics[width= \linewidth]{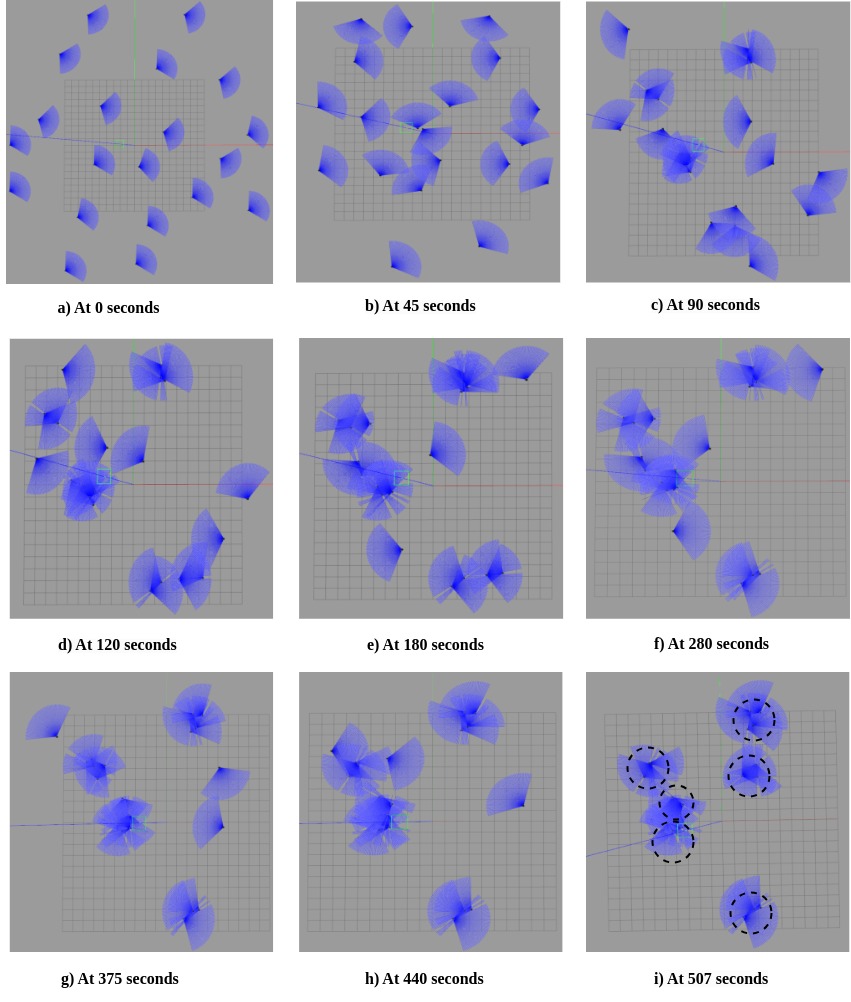}
    \caption{Formation of swarm clusters for swarm size of 20, where blue cones represent the conic FoV of the robots. The parameters considered are $v_{max} = 0.22$~m/s, $a = 0.3$, $\beta = 0.3$, $b = 0.866$, $D_g = 0.875$~m, $\delta_c = 60^\circ$, $D_s = 0.775$~m, $\lambda_i = 10^{-5}$, $R = 3.5$~m.}
    \label{Simulation}
\end{figure*}

We performed various simulations for swarm of robots at different positions and orientations. One such simulation result with sensing range $R =3.5\, \text{m}$ and FoV of $120^{\circ}$ is shown in Figure~\ref{Simulation}. 

The environment size considered for the initialization of all robots is $40\,\text{m}\times 40\,\text{m}$, with the boundary limits are taken as $24\,\text{m} \times 24\,\text{m}$ for the swarm size of 20 ($S = 20$). The boundary limits are local to each robot and used in random goal allocation step. With the minimum cluster size of 3 ($M=3$), the maximum number of clusters that can be formed is 6. In Figure~\ref{Simulation}, we observe that all the robots have converged within 507 seconds forming six clusters, which are indicated  by black dashed circles in Figure~\ref{Simulation}(i). The physical experiments were performed on the Turtle-bot3 burger robot, in presence of real world factors like LiDAR noise (variance), Turtlebot3 speed limits, environmental bounds with lack of communication and GPS. We considered a limited sensing range of $1.6\,\text{m}$ and the experimental area considered was $25~\text{m}^2$. The method efficiently forms clusters without requiring communication or global positioning. Further, we analyzed the success rate of cluster formation for different swarm sizes by varying robot's sensing range and boundary limits for goal update. We evaluated the proposed method across multiple scenarios in both simulation and physical experiments to assess its robustness under varied conditions.

\begin{figure}[!h]
    \centering
    \includegraphics[width=\linewidth]{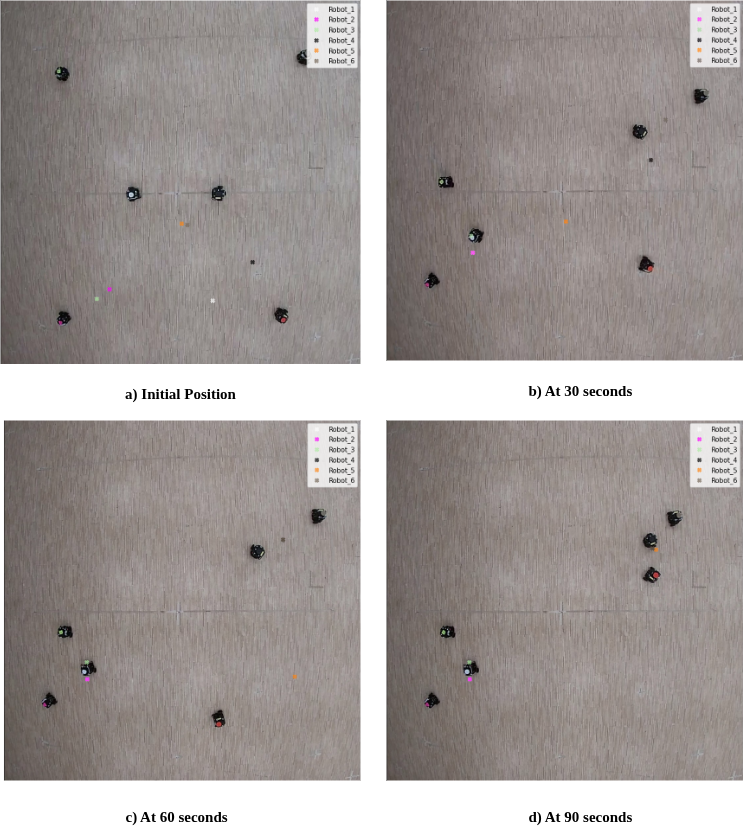}
    \caption{Experimental results with $5\,\text{m}\times 5\,\text{m}$ goal boundary limits and sensing range $R= 1.6\,\text{m}$ for $S= 6$. The parameters considered are $v_{max} = 0.22$~m/s, $a = 0.3$, $\beta = 0.3$, $b = 0.866$, $D_g = 0.5$~m, $\delta_c = 60^\circ$, $D_s = 0.4$~m, $\lambda_i = 10^{-5}$.}
    \label{Exp}
\end{figure}

\subsubsection{Scenario 1: Swarm of robots facing outwards}
This scenario considers an initial configuration where no robot is within the FoV of any other robot. This configuration delays cluster formation instantly after initialization, resulting in the largest time to form cluster(s). A swarm of six robots are placed at random positions in the experimental area, but their orientations faced outwards as shown in Figure \ref{Exp}(a). We observe that two clusters with three members are successfully formed, and none of the robots collide while forming the cluster or within the cluster, maintaining safe distance. We consistently observed successful cluster formation across a wide range of initial robot positions for this scenario.

\subsubsection{Scenario 2: Adaptive Behavior of swarm clusters}\label{adaptive}
\begin{figure}
    \centering
    \includegraphics[width=\linewidth]{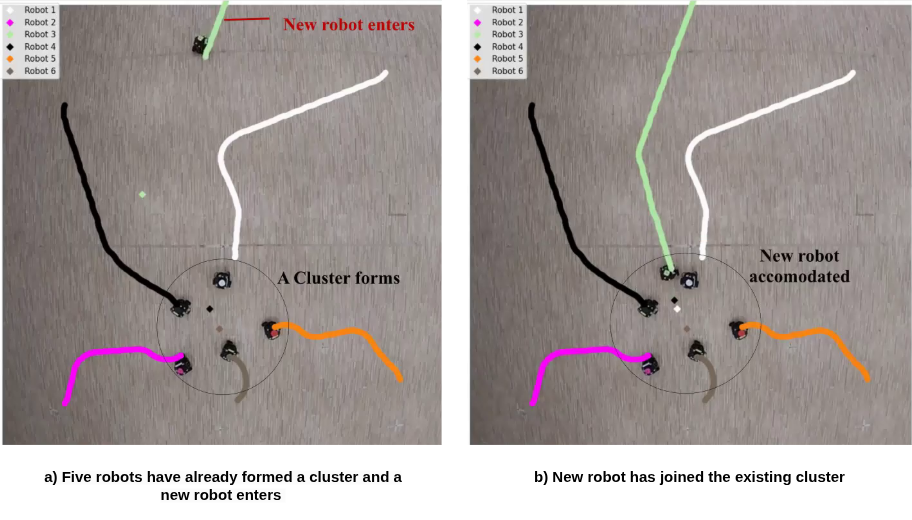}
    \caption{Addition of a new robot into already formed cluster. The parameters considered are $v_{max} = 0.22$~m/s, $a = 0.3$, $\beta = 0.3$, $b = 0.866$, $D_g = 0.5$~m, $\delta_c = 60^\circ$, $D_s = 0.4$~m, $\lambda_i = 10^{-5}$, $R = 1.6$~m.}
    \label{Scene2}
\end{figure}

We tested the method in an additional scenario to assess its adaptive capabilities as part of the emergent behavior. This scenario specifically tests whether an existing cluster can adapt to a newly introduced robot upon its entry into the environment. We randomly placed five robots in the experimental area, which eventually converged to form a single cluster. After cluster formation, a new robot is introduced into the environment. It encounters the existing cluster and efficiently gets accommodated in the same cluster as shown in Figure \ref{Scene2}. This experiment was tested by trying out different initial positions of the new robot as shown in Figure \ref{Scene2_2} (see Appendix). The adaptive behavior of the proposed method has been observed successfully.

\begin{figure*}[!h]
    \centering
    \includegraphics[width=0.8\linewidth]{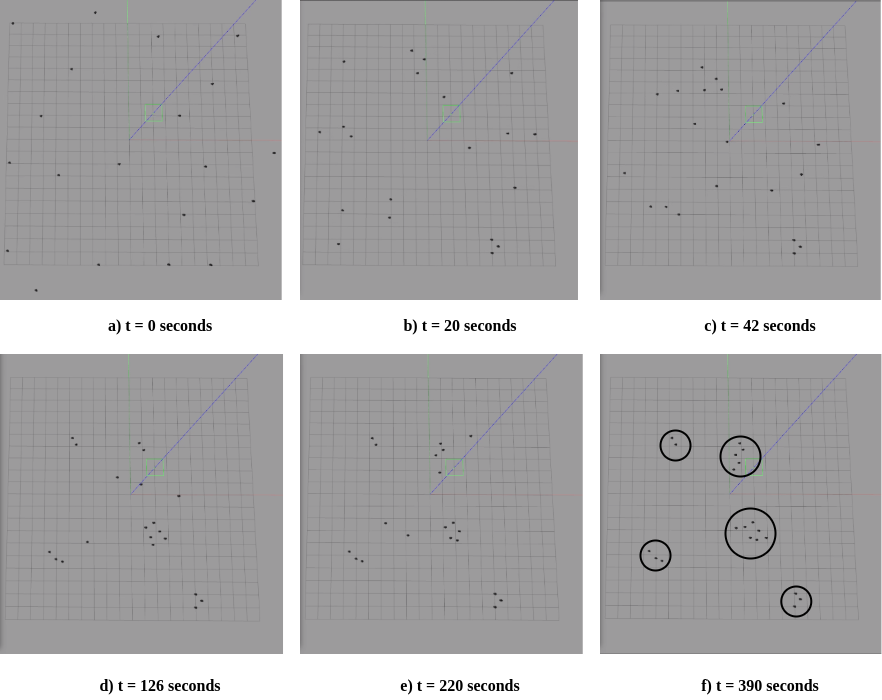}
    \caption{Swarm cluster formation with  circular boundary limits of radius 20 meters. The parameters considered are $v_{max} = 0.22$~m/s, $a = 0.3$, $\beta = 0.3$, $b = 0.866$, $D_g = 0.875$~m, $\delta_c = 60^\circ$, $D_s = 0.775$~m, $\lambda_i = 10^{-5}$, $R = 3.5$~m, Environment area = 40 m x 40 m.}
    \label{Scenario3}
\end{figure*}
\subsubsection{Scenario 3: Self-clustering with local boundaries}
This scenario demonstrates the algorithm’s capability to achieve clustering while each robot operates within its own local environment boundary. Such conditions closely resemble emergency evacuation or underwater exploration scenarios, where global positional information is unavailable. Each robot is assigned a local circular boundary centered at its initial position with a specified radius. The local boundary radius was selected as a design parameter, determined by factors such as the initial robot distribution, swarm size, and total environment area. 

\begin{figure*}[!h]
    \centering
    \includegraphics[width=0.9\linewidth]{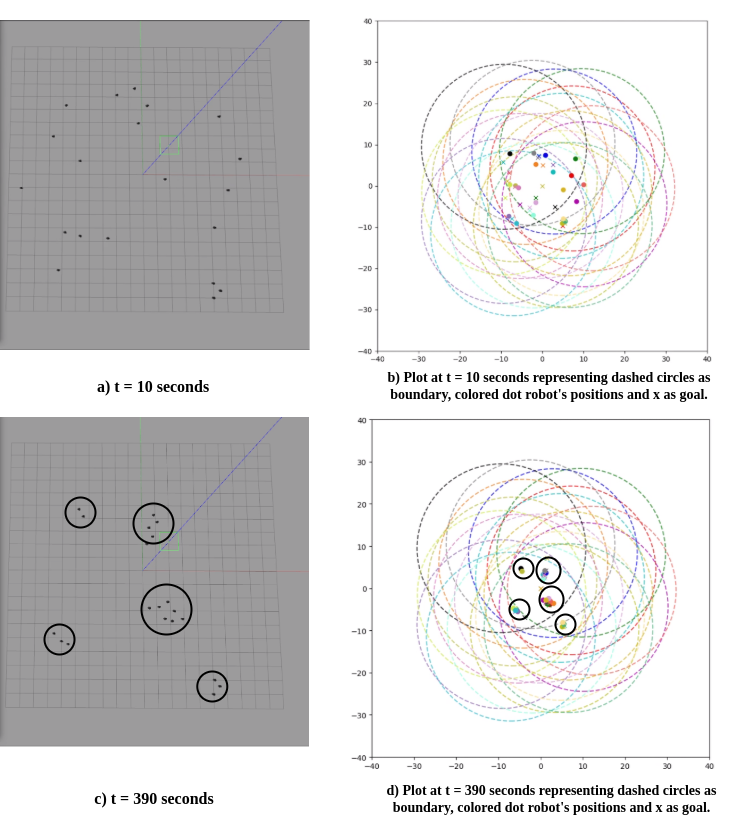}
    \caption{Plot showing the initial and final instants of the simulation, right image show robots' location with respect to the circular boundaries of different robots.}
    \label{Bound plot}
\end{figure*}

A 20 m radius with the selected initial conditions ensured more than 90\%  overlap area providing higher (more than 0.9) probability of encountering other robots to form clusters. Selecting an appropriate boundary radius is crucial to ensure convergence; in this work, a radius of 20 m was determined empirically through a trial-and-error approach see Appendix). Figure~\ref{Scenario3} illustrates the robots’ behavior under this configuration, where the swarm successfully forms compact clusters with $M = 2$ (default setting). The circular boundary visualization, along with the robots’ instantaneous and goal positions, is shown in Figure~\ref{Bound plot}. The two subplots correspond to real-time snapshots from the Gazebo simulation, demonstrating that all robot trajectories remain strictly confined within their respective local boundaries while achieving self-organized clustering. The detailed video showcasing experimental and simulation results is available at \href{https://youtu.be/FQB9HQR9eLE}{this link}. Additional scenario where the global positions are known is presented in Appendix. 

\subsection{Comparative Analysis}
The following quantitative metrics were employed to assess clustering performance across the different swarm clustering algorithms.

\begin{itemize}
    \item \textbf{Compactness} measures the average distance among the robots in a cluster, indicating how tightly the robots are grouped within each cluster.
    \item \textbf{Cohesion} quantifies the overall tendency of robots to move toward the cluster centroid, reflecting group integrity during clustering.
    \item \textbf{Dispersion} evaluates the spatial spread of clusters in the environment, with lower values corresponding to more localized formations.
    \item \textbf{Coverage} represents how effectively the swarm occupies the environment given by the ratio of the total area covered by all robots’ sensing ranges to the environment area.
    \item \textbf{Silhouette score}~\cite{rousseeuw1987silhouettes} assesses the quality of clustering by comparing intra-cluster similarity to inter-cluster separation, where values closer to~1 indicate well-defined clusters.
    \item \textbf{Dunn index}~\cite{dunn1974well} provides a ratio between the minimum inter-cluster distance and the maximum intra-cluster diameter, with higher values denoting better cluster separation.
    \item \textbf{Convergence time} \cite{arvin2011implementation} denotes the total duration required for the swarm to form stable clusters.
    \item \textbf{Clustering Efficiency} metric, adapted from computational performance analysis~\cite{xu2005scaling,pham2020performance}, is defined as the ratio of silhouette score to convergence time, reflecting the trade-off between clustering quality and temporal efficiency.
    \item \textbf{Execution time} is defined as the wall-clock CPU time required to execute a single control–decision cycle of a robot.
    \item The \textbf{Computational Complexity} of a method is derived based on the local operations performed by each robot during clustering.
\end{itemize}

The descriptions of these metrics are further detailed in the Appendix. 
This section presents a performance comparison with existing state-of-the-art methods, namely Particle Swarm Optimization (PSO) algorithm \cite{zheng2023particle}, extended Boids algorithm \cite{HENGSTEBECK2024100160}, and improved Beeclust algorithm \cite{rios2025robot}. The robot and environmental parameters are kept same across all experiments, including initial conditions, sensing range, and field of view (FoV) with the shared objective of achieving swarm self-clustering. The algorithms are implemented using the Robot Operating System (ROS), and physics based simulations are carried out using the ROS-Gazebo environment. The robot used is Turtlebot3, configured with a communication and sensing range of 3.5 meters and a limited FoV of $120^\circ$. The proposed method implementation does not use the communication module. 

The Particle Swarm Optimization (PSO) method~\cite{zheng2023particle} relies on global best position updates, making it sensitive to initial robot placement and dependent on global localization. This dependency often leads to rapid but biased convergence, as robots tend to aggregate at the origin, which acts as the global best position in each run. Although the PSO-based method originates from optimization-driven motion planning, it is adapted here as a baseline to evaluate how optimization-based collective behaviors compare with decentralized local-interaction rules in achieving swarm clustering.
The BeeClust algorithm~\cite{rios2025robot} employs a synthetic temperature field to regulate stop-time behavior but is affected by random walk initialization and density-dependent clustering, resulting in slower and less consistent convergence, particularly in larger swarms. The modified Boids framework~\cite{HENGSTEBECK2024100160} facilitates clustering through consensus-based updates of separation, cohesion, and alignment vectors. Although its behavior resembles that of the proposed method, the latter exhibits superior robustness under varying initial conditions and sensing limitations. In contrast to purely consensus-based methods such as Boids and PSO, the proposed approach introduces an alternating goal assignment mechanism that augments the consensus process. This extension provides key advantages, including low computational time along with highest clustering efficiency, and the emergence of untraceable cluster formations, as validated by the results in Table~\ref{tab:metrics_comparison}. The proposed method is independent of the initial swarm positions because of the proposed method’s distributed initial random-goal assignment, which prevent early bias toward any specific spatial arrangement.

Each robot in the proposed method performs local operations; neighbor detection, goal reassignment, and convergence checks—based solely on local sensing within a fixed range. These tasks execute independently without inter-robot communication or global positioning. The consensus-based goal update rule utilizes the positions of locally detected neighbors, the number of such neighbors is inherently bounded. Each robot is equipped with a finite sensing range $R$, a limited FoV $2\Delta$, and enforces a minimum inter-robot separation $D_s$. These physical and sensing constraints impose an upper bound on the maximum number of robots that can be simultaneously detected within a robot’s sensing region. As a result, the per-robot computation associated with neighbor processing remains constant, and the overall per-robot computational complexity is effectively $\mathcal{O}(1)$, resulting in overall system complexity of $\mathcal{O}(S)$ for swarm size $S$. The improved BeeClust algorithm also scales linearly, $\mathcal{O}(S)$, as it depends only on local field sensing. The extended Boids model, using neighborhood-based alignment and cohesion, exhibits $\mathcal{O}(S \cdot |N|)$ complexity, where $|N|$ is neighborhood size. PSO-APF~\cite{zheng2023particle}, though decentralized, requires global best propagation and particle ranking, resulting in $\mathcal{O}(S^2)$ complexity. 

Table \ref{tab:metrics_comparison} compares the clustering performance of four methods—PSO, Beeclust, Boids, and the proposed approach—over 30 independent trials using both quantitative and qualitative clustering metrics. The reported values represent the mean and corresponding 95\% confidence interval (CI) for each quantitative metric. While PSO and Boids exhibit numerically superior values in certain metrics, these results are primarily driven by centralized attraction behaviors—such as bias toward a known global position (the origin) or virtual “ghost” agents—rather than by adherence to the desired inter-robot spacing constraints. In contrast, the proposed method consistently maintains inter-agent separation strictly within the predefined minimum threshold values \{$D_s, D_g$ = 0.775,0.875\}, resulting in average compactness and cohesion metrics that remain closest to these design thresholds.

\begin{table*}[!h]
\centering
\caption{Comparison of Clustering Quality Metrics Across Local-only Methods for 30 trials each, where PSO and Boids superior performance in some metrics is driven by prioritizing formation of single cluster at origin at each trial. The parameters considered are $v_{max} = 0.22$~m/s, $a = 0.3$, $\beta = 0.3$, $b = 0.866$, $D_g = 0.875$~m, $\delta_c = 60^\circ$, $D_s = 0.775$~m, $\lambda_i = 10^{-5}$, $R = 3.5$~m, and Environment area = 12 m x 12 m. Values highlighted in green represent superior performance within the minimum threshold.}
\label{tab:metrics_comparison}
\scalebox{0.70}{
\begin{tabular}{|l|c|c|c|c|}
\hline
\textbf{Metric} & \textbf{PSO \cite{zheng2023particle}} & \textbf{BeeClust~\cite{rios2025robot}} & \textbf{Boids~\cite{HENGSTEBECK2024100160}} & \textbf{Proposed Method} \\ 
\hline
\textbf{Avg. Compactness} & 2.0489 $\pm$ 0.5174 & \cellcolor{green!15}1.2857 $\pm$ 0.3549 & 0.7014 $\pm$ 0.0371 & \cellcolor{green!15}1.6703 $\pm$ 0.1198 \\ 
\hline
\textbf{Avg. Cohesion} & 1.2086 $\pm$ 0.3445 & 0.7334 $\pm$ 0.2317 & 0.4609 $\pm$ 0.0301 &\cellcolor{green!15} 1.0090 $\pm$ 0.0911 \\ 
\hline
\textbf{Avg. Dispersion} & \cellcolor{green!15}0.5847 $\pm$ 0.2339 & 0.3167 $\pm$ 0.1798 &0.3257 $\pm$ 0.0292 & \cellcolor{green!15}0.5130 $\pm$ 0.1079 \\ 
\hline
\textbf{Avg. Coverage} & \cellcolor{green!15}5.0311 $\pm$ 3.8821 & 1.6521 $\pm$ 2.4053 & 0.4888 $\pm$ 0.0665 & 1.7967 $\pm$ 0.4619 \\ 
\hline
\textbf{Silhouette Coefficient} & 0.2359 $\pm$ 0.0642 & 0.5136 $\pm$ 0.0694 & \cellcolor{green!15}0.8311 $\pm$ 0.0034 & 0.6930 $\pm$ 0.0402 \\ 
\hline
\textbf{Dunn Index} & 1.4434 $\pm$ 0.7634 & 2.5292 $\pm$ 1.0740 & \cellcolor{green!15}3.3196 $\pm$ 0.2394 & 1.6542 $\pm$ 0.3019 \\ 
\hline
\textbf{Convergence Time (s)} & \cellcolor{green!15}35.48 $\pm$ 2.63 & 336.13 $\pm$ 53.85 & 437.44 $\pm$ 138.90 & 96.13 $\pm$ 16.21 \\ 
\hline
\textbf{Clustering Efficiency} & 0.0066 &  0.0015 & 0.0018 & \cellcolor{green!15} 0.0072\\
\hline
\textcolor{blue}{\textbf{Execution Time (in ms)}} & 0.23 & 19.874 & 12.03 &  \cellcolor{green!15} 0.162\\
\hline
\textbf{Computational Complexity} & $\mathcal{O}(S^2)$ &  \cellcolor{green!15} $\mathcal{O}(S)$ & $\mathcal{O}(S.|N|)$ & \cellcolor{green!15} $\mathcal{O}(S)$\\
\hline
\textbf{No. of Clusters Known} & Yes &  \cellcolor{green!15}No & \cellcolor{green!15}No & \cellcolor{green!15} No\\
\hline
\textbf{Untraceability} & No & \cellcolor{green!15} Yes & No & \cellcolor{green!15} Yes\\
\hline
\textbf{Initial Condition Independence} & No & No & \cellcolor{green!15} Yes & \cellcolor{green!15} Yes\\
\hline
\end{tabular}}
\end{table*}

\begin{figure*}[!h]
  \centering
  \begin{subfigure}[t]{0.48\textwidth}
    \centering
    \includegraphics[width=1.1\linewidth]{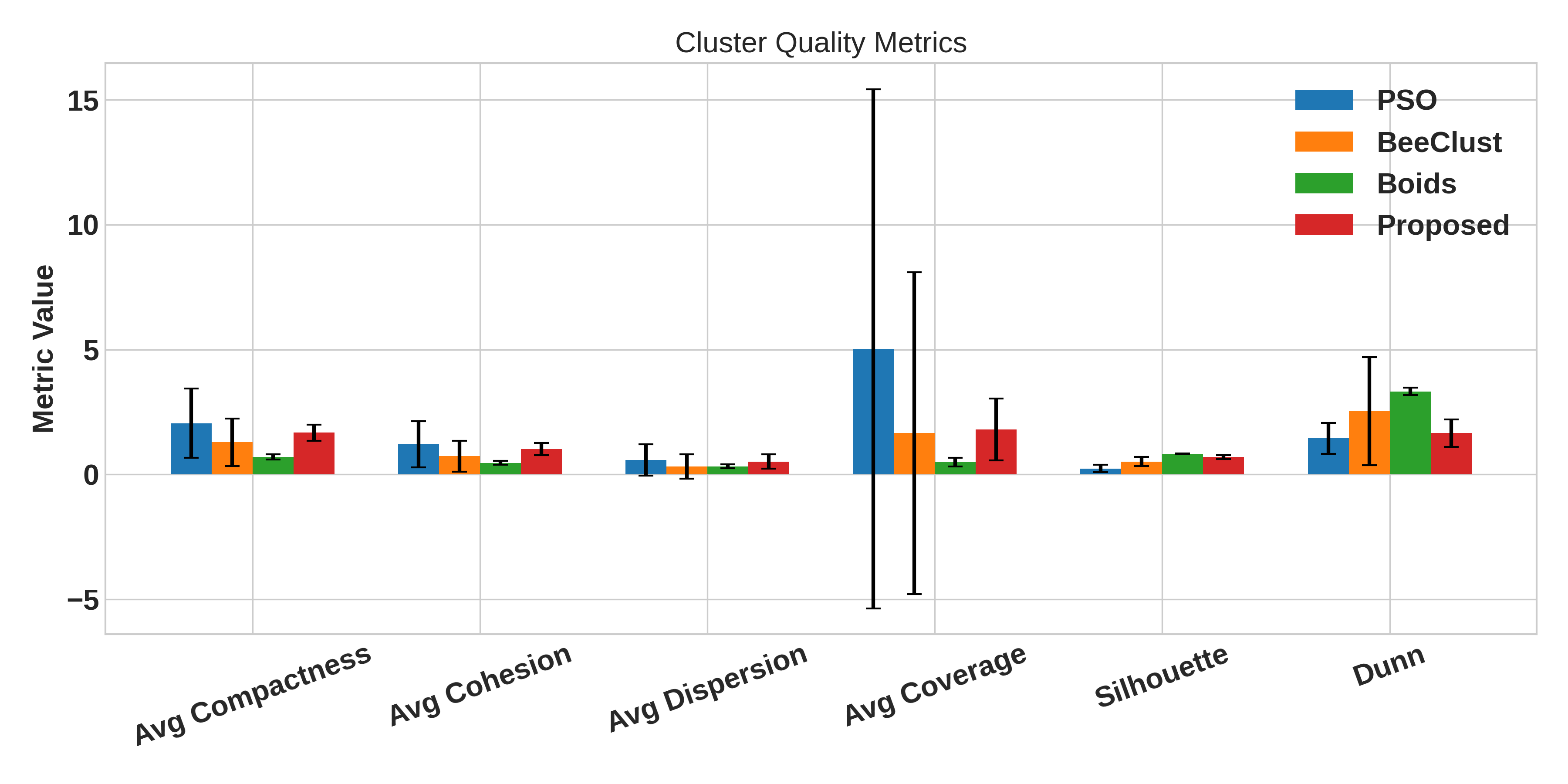}
    \caption{Quality metrics across clustering methods.}
    \label{fig:quality}
  \end{subfigure}
  \hfill
  \begin{subfigure}[t]{0.48\textwidth}
    \centering
    \includegraphics[width=0.9\linewidth]{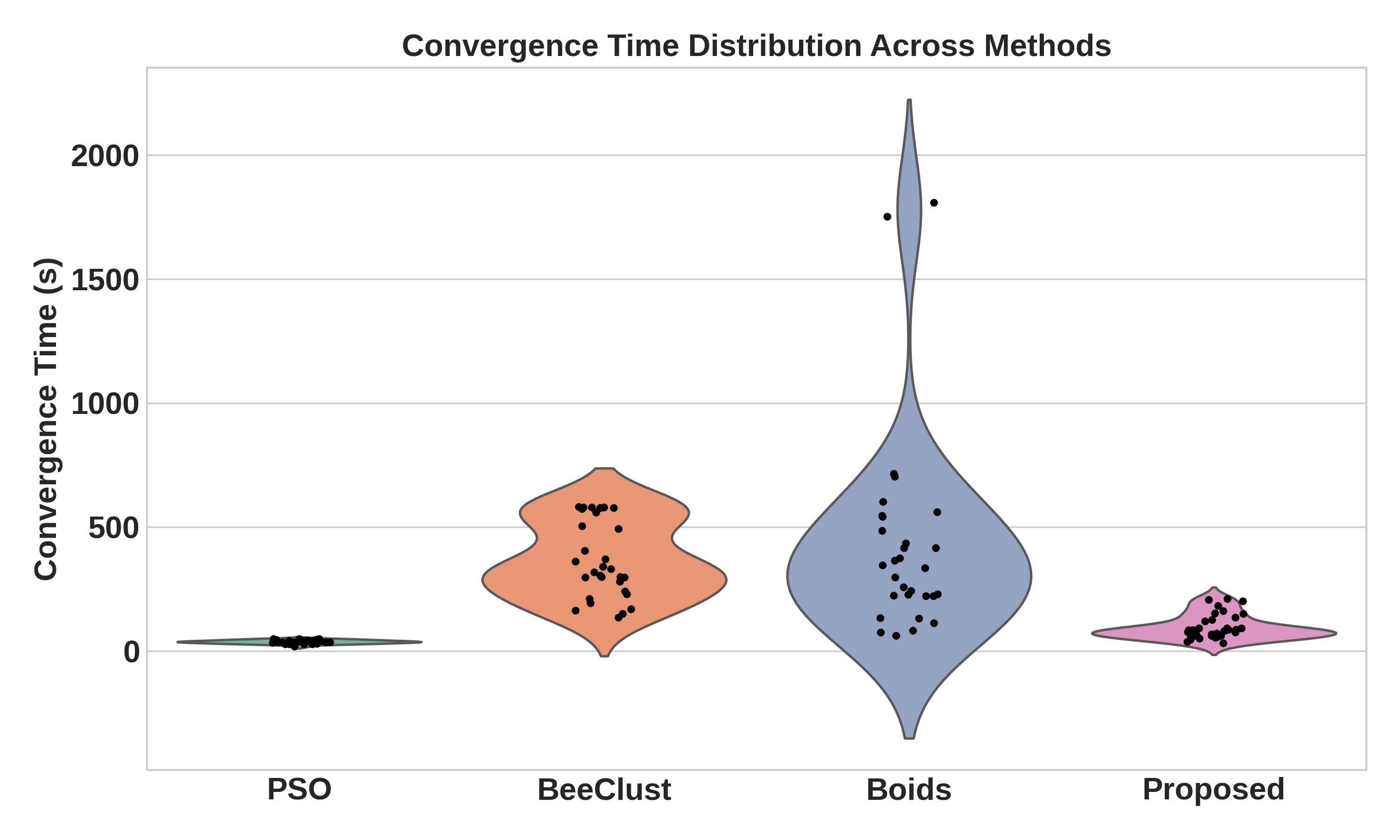}
    \caption{Distribution of Convergence Time (30 trials).}
    \label{fig:violin}
  \end{subfigure}

  \vspace{1.5em} 

  \begin{subfigure}[b]{0.51\textwidth}
    \centering
    \includegraphics[width=\linewidth]{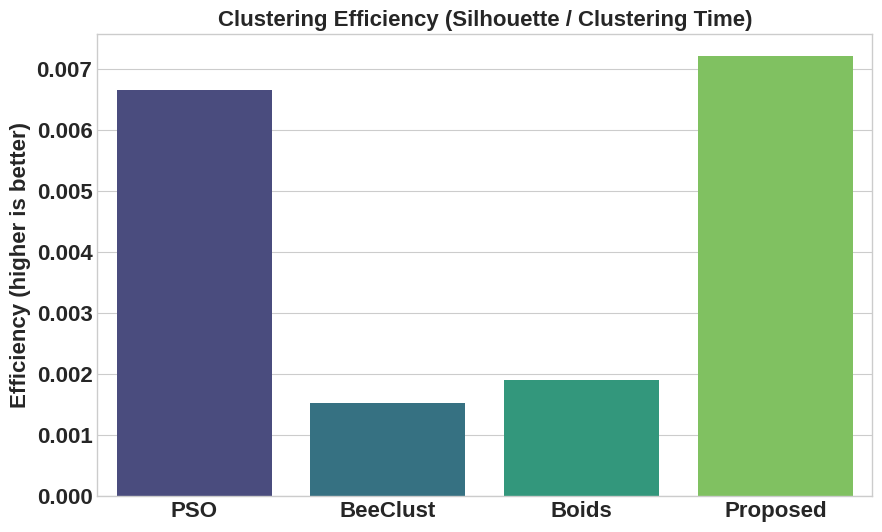}
    \caption{Clustering Efficiency (Silhouette/Time).}
    \label{fig:efficiency}
  \end{subfigure}
  \hfill
  \begin{subfigure}[b]{0.48\textwidth}
    \centering
    \includegraphics[width=0.9\linewidth]{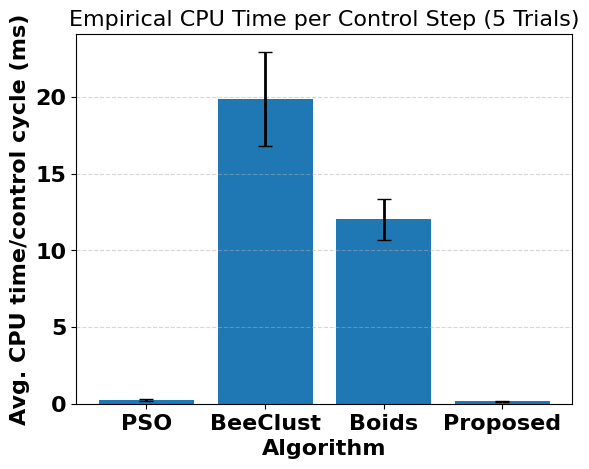}
    \caption{Average CPU time/control cycle.}
    \label{fig:complexity}
  \end{subfigure}

  \caption{Comprehensive comparison of swarm clustering methods. The parameters considered are $v_{max} = 0.22$~m/s, $a = 0.3$, $\beta = 0.3$, $b = 0.866$, $D_g = 0.875$~m, $\delta_c = 60^\circ$, $D_s = 0.775$~m, $\lambda_i = 10^{-5}$, $R = 3.5$~m, Environment area = 12 m x 12 m (a) Quality metrics (30 trials each), (b) Convergence time distributions (30 trials each), 
  (c) efficiency measure (30 trails each), and (d) Average execution time measured inside the decision-making loop for each algorithm, averaged across robots and 5 independent trials on the same hardware.}
  \label{fig:swarm_summary}
\end{figure*} 
Both average compactness and cohesion are lower-bounded by the collision avoidance safe distance $D_g$ and $D_s$ respectively, which defines the minimum permissible inter-robot spacing. This ensures that clusters remain physically realizable and collision-free. The green-highlighted entries in Table~\ref{tab:metrics_comparison} denote metrics that best satisfy these optimal threshold conditions. The proposed method further demonstrates efficient convergence (96 s) and linear computational complexity $\mathcal{O}(S)$, offering a favorable balance between scalability and clustering accuracy. Overall, when considering both quantitative and qualitative metrics, the proposed method demonstrates superior clustering efficiency and stability compared to the local-only baselines. 

Furthermore, the Figure \ref{fig:swarm_summary} compares the swarm clustering methods across four key performance metrics. Figure \ref{fig:quality} shows that the proposed method achieves a higher silhouette score and a competitive Dunn index than PSO and BeeClust, indicating better intra-cluster cohesion and inter-cluster separation. Figure \ref{fig:violin} depicts convergence time variability over 30 trials, where PSO varies the least because of prioritizing single cluster formation due to initial swarm configuration. While Boids and BeeClust exhibit inconsistent run-times the proposed method's convergence time variability lies within a small band which indicates the robustness and initial configuration independence. The normalized efficiency metric in Figure \ref{fig:efficiency} (Silhouette score to convergence time ratio) highlights the proposed method as the most balanced, achieving strong clustering quality with efficient computation. 

Figure \ref{fig:complexity} reports the average CPU time per control cycle, measured inside the decision-making loop and excluding sensing, communication, and physics simulation overhead. The measurements are obtained for a representative swarm size (S=6) and illustrate relative per-step execution cost across algorithms. The proposed approach consistently exhibits low per-step CPU cost, comparable to PSO and substantially lower than BeeClust and Boids, which incur higher overhead due to repeated neighborhood-based computations. Although the proposed method and BeeClust share the same $\mathcal{O}(S)$ theoretical complexity, Figure \ref{fig:complexity} reveals a substantial difference in per-step CPU execution time. This discrepancy arises as BeeClust relies on continuous probabilistic updates and environmental field evaluations at every control cycle, resulting in higher per-step computational overhead. In contrast, the proposed method employs a lightweight dynamic alternating goal-based logic, leading to significantly lower CPU time per control step.

\subsubsection{Factors Affecting Convergence Time}
\begin{figure*}[htbp]
    \centering
    \includegraphics[width=\linewidth]{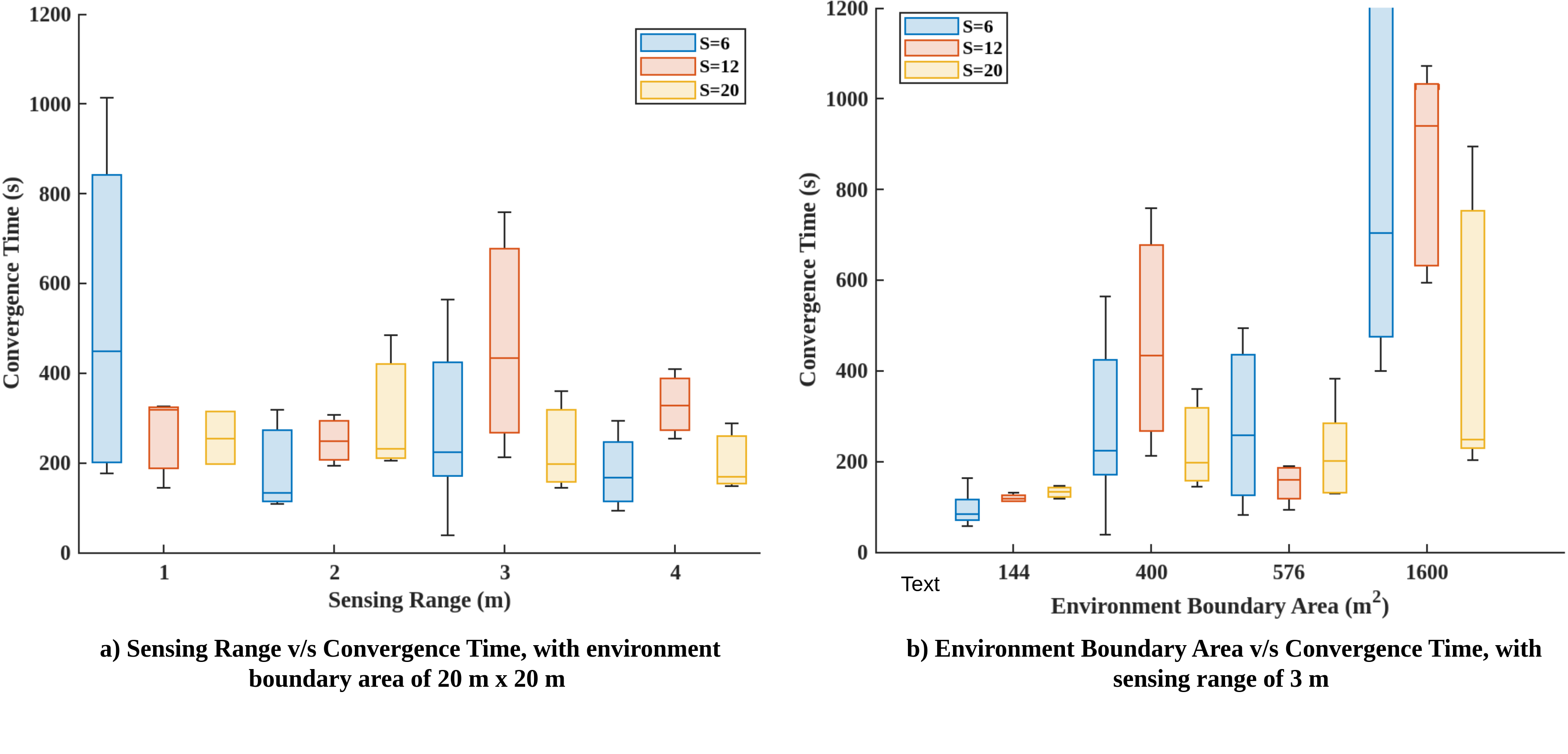}
    \caption{Multiple simulations for each swarm size, showing the median and deviation of obtained values. The parameters considered are $v_{max} = 0.22$~m/s, $a = 0.3$, $\beta = 0.3$, $b = 0.866$, $D_g = 0.875$~m, $\delta_c = 60^\circ$, $D_s = 0.775$~m, $\lambda_i = 10^{-5}$, R$= \{1,2,3,4\}$~m, Environment area = $\{144,400,576,4=1600\}$~m$^2$}
    \label{AR}
\end{figure*}

We show the effect of robot's sensing range and total boundary area for different swarm sizes on the convergence time with more behavioral observations through various simulation and experimental settings.
The factors like sensing range and boundary limits are analyzed to observe their affect on the desired swarm behavior which is quantified by the convergence time. We conducted 20 runs for each factor adjustment to analyze the  swarm self-clustering behavior. The factors included are the sensing range, varying from $1\,\text{m}$ to $4\,\text{m}$, and the boundary limits, ranging from $144\,\text{m}^2$ to $1600\,\text{m}^2$. The results of these simulations are presented in  Figure~\ref{AR}. Both the figures highlight the median and deviation of the values obtained for a change in the factor. In Figure~\ref{AR}(a), all results are obtained by keeping the boundary limits constant while varying the sensing range. Similarly, in Figure~\ref{AR}(b), the results are obtained by fixing the sensing range to 3 meters and varying the boundary limits.

A smaller sensing range was expected to result in higher CT and the smaller boundary limit to result in smaller CT; however, no such pattern is observed. These parameters do not always affect the convergence time linearly because of the dependence on swarm sizes. For instance, in Figure~\ref{AR}(a), the convergence time for the sensing range of 2 \text{m}, is comparatively lesser than that for the sensing range of 3 \text{m}. In Figure~\ref{AR}b), a similar scenario is observed for areas equal to 400 \text{$m^2$} and 576 \text{$m^2$}, where a larger limit resulted in a smaller CT. We also observe that for all the runs, the convergence time is finite which indicates successful convergence of the algorithm. 

\noindent \textbf{Remark:} The convergence time is directly dependent on the swarm size and the environment boundary. This implies that in a sparse arrangement (bigger environment and smaller swarm) the convergence time would increase, reflecting a limitation of the proposed method in achieving rapid clustering.

\subsection{Additional Features}
This section explores various features and functionalities of the proposed algorithm.
\subsubsection{Untraceability}
We analyse the inability to distinguish robot trajectories across different runs within a shared operational environment. In particular, the untraceability is the measure of dissimilarity of a robot's trajectory across runs. To formally evaluate and quantify this property, we adopt two dissimilarity metrics: Dynamic Time Warping (DTW) distance \cite{salvador2007toward} and Jensen–Shannon Divergence (JSD) \cite{menendez1997jensen}.

The DTW distance serves as a trajectory-level dissimilarity measure by computing the minimal cumulative alignment cost between pairs of trajectories, using the physical units of workspace distance (typically meters). Elevated DTW distances, particularly when normalized by the spatial extent of the workspace, indicate that trajectory instances are highly variable and resist deterministic association across trials. In parallel, JSD is employed to compare the empirical probability distributions constructed from trajectory data across different runs. As a bounded, symmetric information-theoretic measure, the JSD quantifies the divergence between these distributions, with higher values reflecting greater unpredictability and diversity in collective motion patterns. By leveraging both DTW and JSD, we provide a rigorous, interpretable framework for evaluating untraceability, capturing both the pairwise dissimilarity of individual paths and the distributional divergence of collective robot behaviors. For reference, the mathematical formulation of these metrics are provided in Appendix~\ref{untrace_ap}.

\begin{figure*}[!h]
    \centering
    \includegraphics[width=\linewidth]{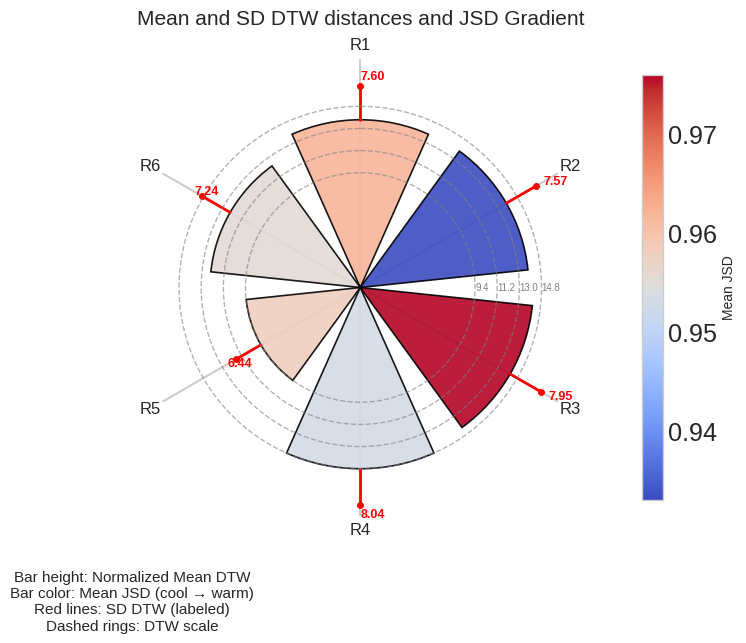}
    \caption{Mean DTW distances and JSD of each robot across 32 identical trails for a swarm size of 6 with  $v_{max} = 0.22$~m/s, $a = 0.3$, $\beta = 0.3$, $b = 0.866$, $D_g = 0.875$~m, $\delta_c = 60^\circ$, $D_s = 0.775$~m, $\lambda_i = 10^{-5}$, $R = 3.5$m, and Environment area = 12m x 12m.  R1, R2,...,R6 represents robot1, robot2,....,robot6 respectively.}
    \label{metrics}
\end{figure*}

The normalized mean, standard deviation intra-DTW distances and mean JSD for each robot across a total of 32 identical runs is shown in Figure~\ref{metrics}. These results provide a quantitative evaluation of trajectory untraceability for a swarm of six robots over 32 identical trials for the swarm size of six. The mean normalized DTW distances for individual robots range from 9.36 to 14.78, with standard deviations between 6.44 and 8.04, indicating substantial variability and dissimilarity among the trajectories generated by each robot across repeated runs—even under identical experimental conditions. This high intra-robot DTW, when normalized by the workspace extent, demonstrates that the spatial paths taken by each robot are highly non-repetitive and resist deterministic association. Complementing this, the mean JSD values, which range from 0.93 to 0.97, reflect a high degree of divergence between the empirical spatial distributions of each robot’s trajectories across trials. These elevated JSD values confirm that the distributional patterns of robot motion are unpredictable and diverse, further supporting the claim of untraceability. Together, these metrics substantiate that the robots’ trajectories are both individually variable and collectively unpredictable. Further supporting analysis is provided in the Appendix.

\begin{figure}[htbp]
    \centering
    \includegraphics[width=\linewidth]{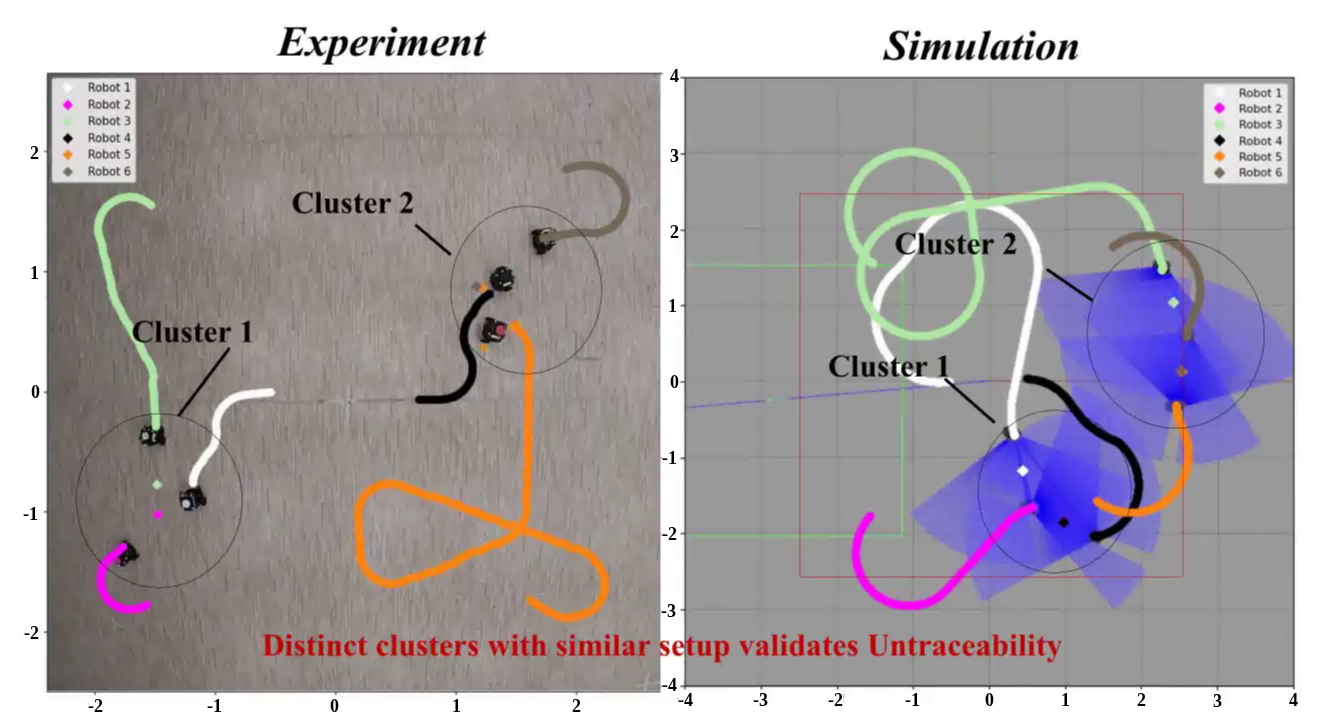}
    \caption{Experiment v/s Simulation Cluster Formation. The parameters considered are $v_{max} = 0.22$~m/s, $a = 0.3$, $\beta = 0.3$, $b = 0.866$, $D_g = 0.5$~m, $\delta_c = 60^\circ$, $D_s = 0.4$~m, $\lambda_i = 10^{-5}$, $R = 1.6$~m.}
    \label{CompES}
\end{figure}

We also examined untraceability across experimental and simulation settings under identical initial conditions, environment, sensing parameters, and swarm size. In the experimental setup, the swarm achieved convergence in approximately 90 seconds, forming two clusters. The corresponding simulation under identical conditions exhibited a slightly faster convergence time of 80 seconds, also resulting in two clusters. This close agreement between experiment and simulation demonstrates the consistency and reliability of the proposed clustering method across both real-world and simulated environments. However, cluster formations vary in location and number of members, while robot trajectories remain untraceable in simulations, as shown in Figure~\ref{CompES}.

\subsubsection{Scalability} 
The scalability of the proposed method was evaluated across varying swarm sizes. Simulations were conducted using the Robot Operating System (ROS Noetic) integrated with Gazebo on Ubuntu 20.04 LTS. Convergence performance was analyzed for swarm sizes of 
(S $= 6,12,20$), with Figure~\ref{AR} illustrating the convergence behavior for a swarm of 20 robots. For larger swarm sizes (S $= 50,75,100$), the implementation was transitioned to MATLAB due to computational constraints encountered in handling large-scale swarms within the physics-based Gazebo simulator.

\begin{figure*}[htbp]
  \centering
  \begin{subfigure}[t]{0.48\textwidth}
    \centering
   \includegraphics[width=\linewidth]{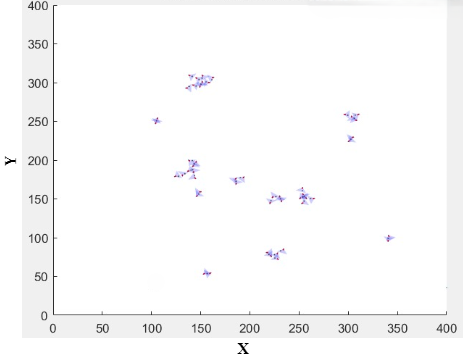}
    \caption{Swarm Size = 50 robots}
    \label{fig:scale1}
  \end{subfigure}
  \hfill
  \begin{subfigure}[t]{0.48\textwidth}
    \centering
    \includegraphics[width=1.06\linewidth]{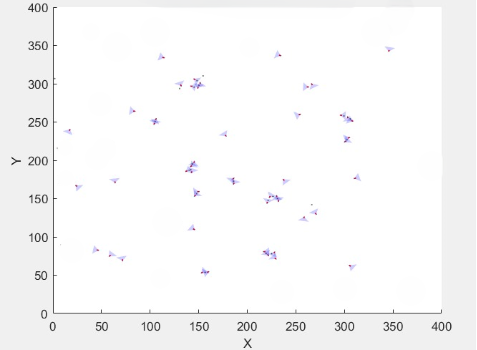}
    \caption{Swarm Size = 100 robots}
    \label{fig:scale2}
  \end{subfigure}
  \caption{Convergence of larger swarms using MATLAB simulations}
  \label{Scalability}
\end{figure*}

The method consistently achieved formation of stable swarm clusters, even at larger swarm sizes. In particular, the method allows for the selection of any swarm size since each robot has complete autonomy; its computations are independent and remain unaffected by the total number of robots in the swarm. Though the convergence time will vary with the change in swarm size. Figure \ref{Scalability} shows the plot of the robots' final positions on a cartesian plane after applying the proposed method for large swarm sizes. The results confirm the method’s effectiveness in maintaining clustering behavior as the swarm size increases.

\subsubsection{Minimum Cluster Size}
The swarm self-clustering involves formation of swarm clusters and it is thus important to assess the number of clusters formed. The upper bound on number of clusters formed in the proposed method is given by \eqref{upperbound}. We varied the minimum cluster size parameter M to explore the formation of different numbers of clusters. We conducted 20 simulation runs across various swarm sizes to evaluate the clustering behavior under different conditions. The experiments included scenarios with and without a constraint on the minimum cluster size. For instance, with a swarm size of 20 (see Figure~\ref{Simulation}), setting the minimum cluster size $M = 3$ imposes an upper bound of six on the number of clusters that can be formed. In contrast, when no lower bound is enforced (effectively $M = 2$), the average number of clusters observed increases to approximately ten. Table~\ref{bound} presents a comparative assessment of the number of clusters formed under both conditions. Results indicate that enforcing a minimum cluster size reduces the maximum number of clusters, albeit at the cost of increased convergence time (CT). Furthermore, when $M \geq S/2 + 1$, the method consistently leads to the formation of a single cluster. While the method showcases expected asymptotic convergence in such cases, there remains significant scope for optimizing convergence time, as discussed in the future work.

\begin{table}
\begin{center}
\centering \caption{Effect of minimum cluster size on number of clusters (20 samples) for a swarm size of 20 robots}
\label{bound}
\begin{tabular}{|p{2cm}|p{3cm}|p{5cm}|}
 \hline
 \textbf{Criteria} & \textbf{No. of clusters (mean)} &  \textbf{Mean Convergence Time (in sec)}\\  
 \hline
 No limit & 10 & 120 \\
 $M=3$  & 6 & 500 \\
 $M=11$ & 1 & 1200 for $20\%$ runs\\
 \hline
\end{tabular}
\end{center}
\end{table}

\section{Conclusions and Future Work}\label{C}
This study presents a novel swarm-based self-clustering framework in which range-sensing robots autonomously organize into spatially coherent groups without any communication, global positioning, or centralized control. The key technical innovation lies in a mechanism that enables each robot to dynamically alternate between either cooperative consensus or random goal exploration based solely on its local neighborhood size. This ensures continuous balance between exploration and convergence, while a distributed termination rule allows clusters to stabilize autonomously once local equilibrium is achieved—without requiring global synchronization or explicit communication.

The proposed method demonstrates empirical convergence across more than 35 independent trials, despite lacking robot identities or prior knowledge of cluster parameters such as the number, size, or composition of groups. It achieves linear computational complexity, ensuring scalability and efficiency for large decentralized swarms operating in resource-constrained environments. Extensive evaluations confirm the approach’s robustness under sensor noise, adaptability to dynamic swarm membership, and ability to form compact, collision-free clusters with untraceable robot trajectories—a property rarely observed in local-only swarm algorithms.

The environment boundaries of each robot support the requirements of scenarios from underwater swarm collection and behavior for grouping under emergency evacuation. Moreover, an important observation is that knowledge of the common target location (i.e., the meeting point of the robots) compromises the untraceability of the swarm. While the current study addresses cluster formation, the underlying approach can be extended to broader swarm behaviors such as aggregation and formation, particularly in communication-constrained environments like underwater domains where global information is unavailable. Enabling complete autonomy at the robot level also offers resilience against communication delays and security threats such as hacking. 

Theoretically, when operating near the boundary condition $d_g=D_g$, oscillatory behavior may occur for fast moving robots. Future extensions may incorporate a hysteresis margin or probabilistic rule to enhance robustness against sensor noise and improve stability. Another future direction is on enhancing the method’s applicability to real-world scenarios by incorporating obstacle-rich environments, where obstacles are detected solely through onboard range sensing. Additionally, improving convergence time for sparse deployments of swarm can be targeted by using memory-based random exploration.

\vspace{0.5cm}
\noindent \textbf{Author Contributions}
Sweksha Jain: Conceptualization, Simulation, Experimentation, Writing (original draft), Editing; Rugved Katole: Experimentation;
Leena Vachhani: Supervision, Editing, Writing

\noindent \textbf{Funding}
No funds, grants or other support were received for this work.

\noindent \textbf{Availability of data and materials}
The link of GitHub repository (limited access) for the data and code is provided in supplementary material.

\section*{Declarations}
\textbf{Acknowledgment}
The authors gratefully acknowledge Dr. Somnath Buriuly for his valuable discussions and constructive feedback that contributed to the refinement of this revised manuscript. The authors also acknowledge the use of OpenAI’s ChatGPT tool for editorial and language clarity improvements. All scientific ideas, experimental designs, analyses, and conclusions presented in this work are solely those of the authors.

\noindent \textbf{Conflict of Interest}
The authors have no conflict of interest in the material discussed in this article.

\noindent \textbf{Consent to publication}
The authors give consent to publish this work.

\bibliographystyle{tfnlm}
\bibliography{root}

\section*{Appendix} \label{Appendix}

\subsection*{Scenario 2: Adaptive behavior}\label{scene2}
\begin{figure*}[!h]
    \centering
    \includegraphics[width=\linewidth]{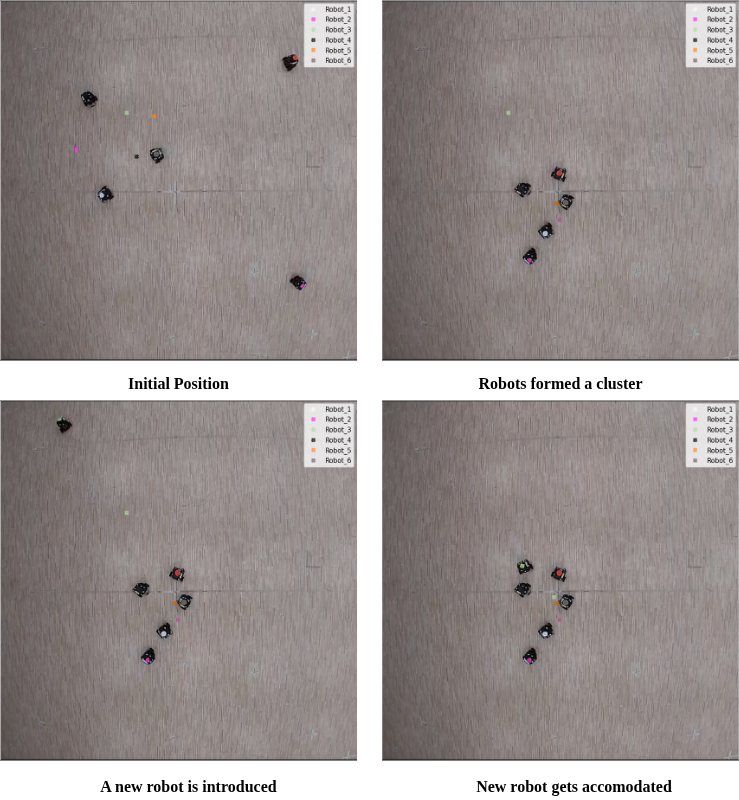}
    \caption{Experiment on adaptive behavior}
    \label{Scene2_2}
\end{figure*}

Figure~\ref{Scene2_2} demonstrates the robustness of the adaptive behavior, showing that a new robot can seamlessly integrate into an existing swarm cluster even when entering from a different position. Specifically, the figure illustrates a new robot joining an already formed cluster from a location distinct from the one presented in Section~\ref{adaptive}.

\subsection*{Scenario 3: Self-clustering with local boundaries}\label{scenario3}
To quantitatively evaluate the effect of the local boundary radius, simulations were conducted for radii of 5 m, 10 m, and 20 m using a swarm of 20 robots. Each configuration was tested in five independent runs with identical initial conditions, and the number of robots converging to clusters was recorded after 7 minutes. For smaller boundaries (5 m and 10 m), complete convergence was rarely achieved (0 \% and 20 \%, respectively) due to limited overlap and low encounter probability. Increasing the radius to 20 m markedly improved performance, achieving full convergence in 60 \% of the runs. Furthermore, extending the random-goal sampling range from [-15 m, 15 m] to [-20 m, 20 m] enabled full convergence in all trials as reported in Table \ref{ablation}. These results confirm that a 20 m local boundary which yields $\approx$ 90\% overlap among robots' local circular boundary, provides an effective balance between exploration and clustering efficiency.

\begin{table}[!h]
\begin{center}
\centering \caption{Ablation Summary for $S=20$ with 5 trials each radii}
\label{ablation}
\begin{tabular}{|p{2cm}|p{3cm}|p{4cm}|p{3.5cm}|}
\hline
\textbf{Radius (m)} & \textbf{Random Goal Limit (m)} &  \textbf{Average Robots Converged} & \textbf{Runs with Full Convergence (\%)}\\  
\hline
5 & [-10,10] & 12-16 & 0\%\\
10 & [-10,10] & 14-16 & 20\% \\
20 & [-15,15] & 18-19 & 60\% \\
20 & [-20,20] & 20 & 100\% \\
\hline
\end{tabular}
\end{center}
\end{table}

\vspace{-1cm}
\subsection*{Scenario 4: Self-clustering with global positioning}\label{scenario4}
To evaluate the effect of global positioning and knowledge of common target on the method's performance, we implemented two variants of the proposed method. In the first variant, each robot has access to the global positions of all the robots in the environment. In the second variant, robots are provided with the exact locations of target clusters. The performance comparison is summarized in Table~\ref{com2}, where reported CT is the average over five independent trials for each case. Interestingly, the convergence time achieved using the knowledge of global positioning case is  comparable to the proposed method, which relies solely on the relative positioning through onboard sensing.

Further, the performance analysis of second variant that considers known target location of clusters with global positioning shows that the clusters are formed quicker than the proposed solution of self-clustering (with range only sensing robots). To evaluate the untraceability feature, we also performed ANOVA tests on the trajectories of individual robots across all trials. 
For the first variant of known global positioning, the results indicated statistically significant differences between runs for both x and y coordinates ($p_x$ = 7.37×$10^{-189}$; $p_y$ = 1.61×$10^{-24}$). These extremely low p-values confirm that the robot’s trajectory is not repeatable across runs, demonstrating the untraceability property of the proposed algorithm, wherein individual robot paths vary randomly despite identical starting conditions. However, for the known target location of clusters, the trajectories are traceable. It is clear that the range-sensing robots forming clusters can converge faster if they cooperate and communicate the target location of clusters. Just by knowing the position of other robots (adding GPS as onboard sensor and sharing the location with nearby robots) may not be very beneficial unless the robots share the intent as well. But, sharing the target location of clusters compromises on the untraceability of the robots' trajectories.

\begin{table}[!h]
\begin{center}
\caption{Performance comparison with known global information  (An average over five trials for a swarm size of six)}
\label{com2}
   \begin{tabular}{|m{7cm}|m{2.75cm}|m{2.5cm}|}
    \hline
    \textbf{Proposed Algorithm} & \textbf{Untraceability} & \textbf{CT mean (s)}\\
    \hline
    \textbf{Proposed Swarm Self-clustering} & Yes & 102\\
    With known global positioning & Yes & 94\\
    With known global positioning and cluster target & No & 77\\
    \hline
   \end{tabular}
\end{center}   
\end{table}

\subsection*{Cluster Quality Metrics} \label{quality}
To evaluate the quality of swarm clustering, six widely used cluster validation metrics are considered: \textit{compactness}, \textit{cohesion}, \textit{coverage}, \textit{dispersion}, \textit{Silhouette coefficient}, and the \textit{Dunn index}. Their definitions and mathematical formulations are as follows.

\paragraph*{1. Compactness ($C_p$):}
Measures the intra-cluster similarity or how tightly robots are grouped within a cluster.
\begin{equation}
C_p = \frac{1}{K} \sum_{i=1}^{K} \frac{1}{n_i} \sum_{j=1}^{n_i} \| x_{ij} - \mu_i \|^2
\end{equation}
where $K$ is the total number of clusters, $n_i$ is the number of robots in cluster $i$, $x_{ij}$ denotes the position of robot $j$ in cluster $i$, and $\mu_i$ is the centroid of cluster $i$.

\paragraph*{2. Cohesion ($C_h$):}
Quantifies the average pairwise distance between robots within the same cluster.
\begin{equation}
C_h = \frac{1}{K} \sum_{i=1}^{K} \frac{2}{n_i(n_i - 1)} \sum_{p<q}^{n_i} \| x_{ip} - x_{iq} \|
\end{equation}

\paragraph*{3. Coverage ($C_v$):}
Indicates how well the entire environment is covered by all clusters.
\begin{equation}
C_v = \frac{A_{\text{covered}}}{A_{\text{total}}}
\end{equation}
where $A_{\text{covered}}$ is the total area occupied by the robots and $A_{\text{total}}$ is the area of the workspace.

\paragraph*{4. Dispersion ($Disp$):}
Represents the separation between clusters, i.e., how far apart cluster centroids are.
\begin{equation}
Disp = \frac{2}{K(K-1)} \sum_{i<j}^{K} \| \mu_i - \mu_j \|
\end{equation}

\paragraph*{5. Silhouette Coefficient ($S_c$):}
Evaluates how well robots fit within their own cluster compared to others.
\begin{equation}
S_c = \frac{1}{N} \sum_{i=1}^{N} \frac{b_i - a_i}{\max(a_i, b_i)}
\end{equation}
where $a_i$ is the mean intra-cluster distance of robot $i$, and $b_i$ is the mean nearest-cluster distance.

\paragraph*{6. Dunn Index ($D_i$):}
Assesses cluster compactness and separation simultaneously.
\begin{equation}
D_i = \frac{\displaystyle \min_{1 \leq i < j \leq K} \delta(C_i, C_j)}{\displaystyle \max_{1 \leq k \leq K} \Delta(C_k)}
\end{equation}
where $\delta(C_i, C_j)$ is the inter-cluster distance between clusters $i$ and $j$, and $\Delta(C_k)$ is the intra-cluster diameter of cluster $k$.

\subsection*{Untraceability}\label{untrace_ap}
In this section, we examine the untraceability feature of swarm self-clustering. Since the goal positions are randomly allocated, the resulting robot trajectories vary across runs, producing non-deterministic final clusters. To evaluate this property, we conducted five trials for each swarm size ${6,12,20}$ without altering initial conditions. Observations included the number of clusters formed, the trajectories of individual robots, and the membership composition of each cluster. Each robot was assigned a unique identifier to facilitate data recording and analysis.

\begin{table}[htbp]
\centering
\caption{Validation for Untraceable clusters}
\label{table:3}
\begin{center}
\scalebox{0.8}{
{
\begin{tabular}{|p{1.2cm}|p{1cm}| p{1.5cm}|p{11cm}| }
 \hline
 \textbf{Swarm Size} & \textbf{Trial No.} & \textbf{No. of clusters} & \textbf{Members of clusters (represented by robot ID)}\\
 \hline 
 \multirow{5}{2cm}{S=6} & 1 & 1 & \{1,2,3,4,5,6\}\\
 & 2 & 2 & \{1,3,4\},\{2,5,6\}\\
 & 3 & 3 & \{1,3\},\{2,4\},\{5,6\}\\
 & 4 & 2 & \{4,5\},\{1,2,3,6\}\\
 & 5 & 2 & \{3,6\},\{1,2,4,5\}\\
 \hline
 \multirow{5}{2cm}{S=12} & 1 & 3 & \{10,3,1,9\}, \{7,8,6,5\}, \{12,4,2,11\}\\
 & 2 & 4 & \{12,3,9,1\}, \{8,10\}, \{7,6\}, \{10,5,2,4\}\\
 & 3 & 3 & \{4,2,10,9,12\}, \{7,8\}, \{6,5\}, \{3,1,11\}\\
 & 4 & 3 & \{1,2,3,9\}, \{8,6,4,7,11\}, \{5,10,12\}\\
 & 5 & 3 & \{6,7,2,11,5,8\}, \{3,1,9\}, \{10,12\}\\
 \hline
 \multirow{5}{2cm}{S=20} & 1 & 5 & \{11,18\},\{5,12,8\}, \{10,7\}, \{14,4,1,6,3,16,17,20,2\}, \{9,19,15,13\}\\
 & 2 & 4 & \{9,13,8,10\}, \{12,2,5,7,11\}, \{4,20,15,16,19,14\}, \{3,18,17,6,1\}\\
 & 3 & 5 & \{2,10,18,20,16,11\}, \{13,15\}, \{7,8,9\}, \{3,17,6,1\}, \{4,5,12,14,19\}\\
 & 4 & 6 & \{4,1\}, \{9,18,6\}, \{12,15,14,19\}, \{7,8,5,10,16,3\}, \{13,20,2\}, \{11,17\}\\
 & 5 & 6 & \{5,14\}, \{7,11,2,18,12,1,19\}, \{4,8\}, \{3,6,17\}, \{16,13\}, \{20,15,9,10\}\\
 \hline
\end{tabular}}}
\end{center}
\end{table}

\begin{figure*}[!h]
    \centering
    \includegraphics[width= \linewidth]{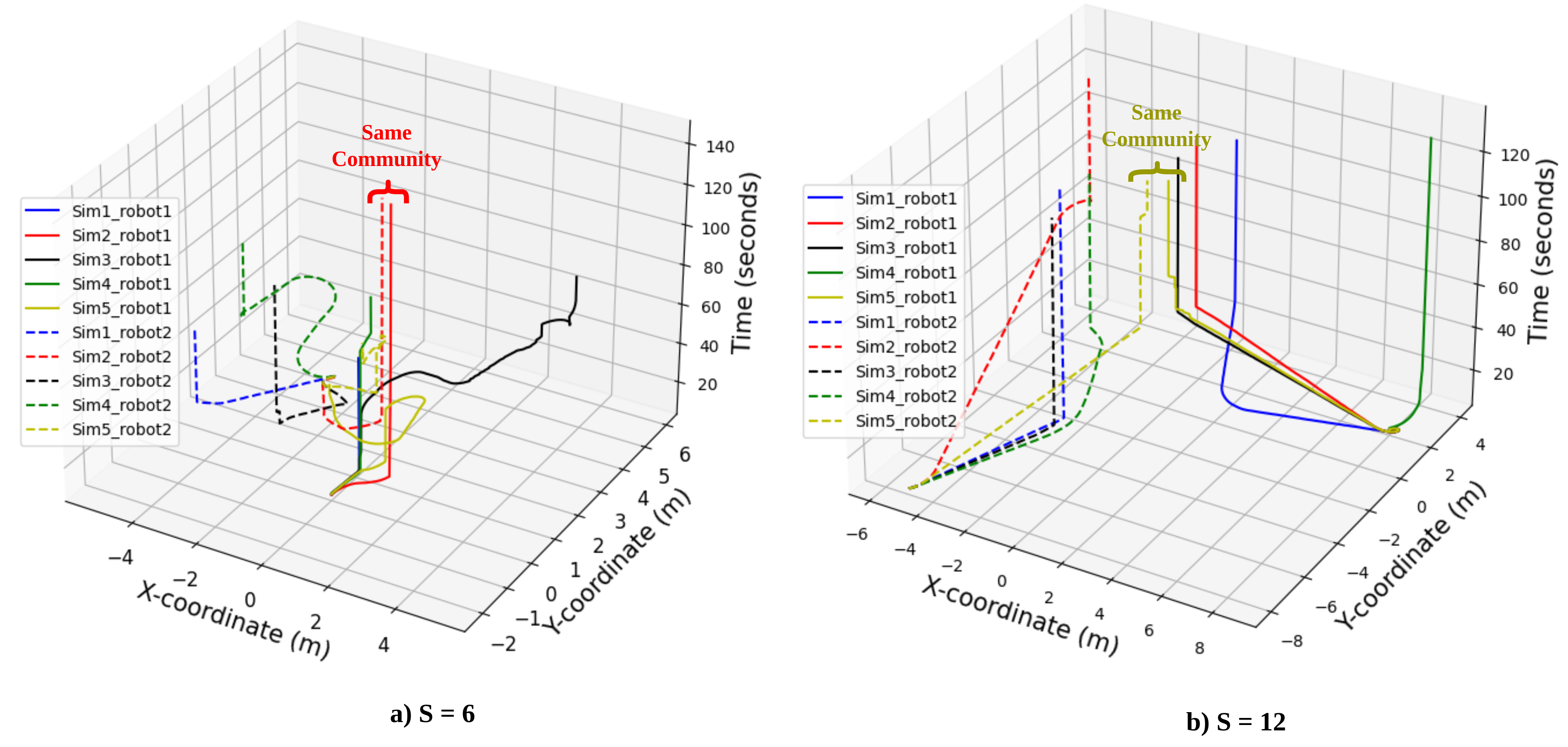}
    \caption{Two robot trajectories for 5 trials showing untraceability of robot trajectories, where S is swarm size}
    \label{untrace}
\end{figure*}

The results, summarized in Table~\ref{table:3}, highlight two levels of variability: (i) for swarm size 20, the number of clusters remained constant across runs but the composition of members within each cluster changed, and (ii) across swarm sizes, the total number of clusters also varied across trials. For illustration, trajectories of two representative robots across five runs are shown in Figure~\ref{untrace}, where the likelihood of the same robots converging into identical clusters across trials is visibly low.

\noindent \textbf{1. Dynamic Time Warping}\\
We employ Dynamic Time Warping (DTW) distances to measure trajectory similarity: higher DTW distances between the same robot’s trajectories across runs imply lower traceability. For two trajectories $X = {x_1, x_2, \dots, x_n}$ and $Y = {y_1, y_2, \dots, y_m}$, the DTW distance is defined as
\begin{equation}
    DTW(X, Y) = \min_{W} \sum_{(i,j) \in W} d(x_i, y_j),
\end{equation}
where $d(x_i, y_j)$ is a local distance measure (e.g., Euclidean), and $W$ is a valid warping path.

\noindent \textbf{2. Jensen Shannon Divergence}\\
The Jensen-Shannon divergence measures the similarity between two probability distributions \( P \) and \( Q \).
It is a symmetric and bounded version of the Kullback--Leibler divergence, where \( M = \frac{1}{2}(P + Q) \) is their average distribution.
A value of \( \mathrm{JSD} = 0 \) indicates identical distributions, while larger values indicate increasing dissimilarity.
\begin{equation}
\mathrm{JSD}(P \parallel Q) = \frac{1}{2} D_{\mathrm{KL}}(P \parallel M) + \frac{1}{2} D_{\mathrm{KL}}(Q \parallel M)
\end{equation}
\begin{equation}
M = \frac{1}{2}(P + Q)
\end{equation}
\begin{equation}
D_{\mathrm{KL}}(P \parallel Q) = \sum_{i} p_i \log \frac{p_i}{q_i}
\end{equation}

\section*{Supplementary Material}
The code and the data can be found on the GitHub repository \href{https://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@github.com/swe-sys/paper-review-repo.git}{repo link}.

\end{document}